\pdfoutput=1
\PassOptionsToPackage{most}{tcolorbox}
\documentclass[11pt]{article}

\usepackage[]{ACL2023}

\usepackage{times}
\usepackage{latexsym}
\usepackage{amsmath}
\usepackage{amssymb}
\usepackage{graphicx}
\usepackage{tcolorbox}
\usepackage{enumitem}
\usepackage{multirow}
\usepackage{tabularx}
\usepackage{booktabs}
\usepackage[most]{tcolorbox}  
\tcbuselibrary{breakable} 
\usepackage{longtable}
\usepackage{xcolor}
\usepackage{listings}
\usepackage{hyperref}
\usepackage[normalem]{ulem}
\useunder{\uline}{\ul}{}
\usepackage[ruled,vlined]{algorithm2e}
\usepackage{fontawesome5} 
\usepackage{hyperref}     

\lstdefinestyle{prompt}{
    basicstyle=\ttfamily\footnotesize,
    backgroundcolor=\color{gray!10},
    frame=single,
    breaklines=true,
    keepspaces=true,
    columns=fullflexible,
    captionpos=b
}

\usepackage[T1]{fontenc}

\usepackage[utf8]{inputenc}

\usepackage{microtype}
\usepackage{natbib}

\usepackage{inconsolata}
\usepackage{amssymb}

\usepackage{amsthm,amsmath,amsfonts,bm,xspace}
\usepackage{color}

\def\eqref#1{equation~\ref{#1}}

\def\1{\bm{1}}

\DeclareMathAlphabet{\mathsfit}{\encodingdefault}{\sfdefault}{m}{sl}
\SetMathAlphabet{\mathsfit}{bold}{\encodingdefault}{\sfdefault}{bx}{n}

\newcommand{\model}[1]{\textsc{#1}\xspace}

\newcommand{\gpto}{\model{GPT-4o}}
\newcommand{\dataset}[1]{\textsc{#1}\xspace}
\newcommand{\ourdata}{\dataset{LIC}}
\newcommand{\labeled}{\dataset{LIC\textsubscript{L}}}
\newcommand{\unlabeled}{\dataset{LIC\textsubscript{U}}}
\newcommand{\method}[1]{\textsc{#1}\xspace}
\newcommand{\ourname}{\method{LePREC}}
\newcommand{\subpara}[1]{%
  \noindent                       
  {\textit{\uline{#1:}}\,}        
}

\newcommand{\rev}[1] {{\textcolor{black}{#1}}}

\newcommand{\find}[1]{%
  \begin{tcolorbox}[
    leftrule=1mm, toprule=0mm, bottomrule=0mm,
    breakable,
    left=1pt,  right=2pt,
    top=0.2em,   bottom=0.2em,   
    boxsep=1pt,              
    fontupper=\small
    ]%
    \em #1%
  \end{tcolorbox}}

\defcitealias{microsoftphi4}{Microsoft Phi-4}
\defcitealias{gptoss20b}{GPT-oss-20B}
\defcitealias{commonwealth_members}{Commonwealth law foundations}
\defcitealias{clj}{Current Law Journal}
\defcitealias{yang2025qwen3}{Qwen3-14B}

\title{\ourname: Reasoning as Classification over Structured Factors for Assessing Relevance of Legal Issues}

\begin{document}

\author{
{\bf Fanyu Wang$^\spadesuit$, Xiaoxi Kang$^\diamondsuit$, Paul Burgess$^\clubsuit$, Aashish Srivastava$^\clubsuit$,} \\
{\bf Chetan Arora$^\spadesuit$, Adnan Trakic$^\heartsuit$, Lay-Ki Soon$^\diamondsuit$, Md Khalid Hossain$^\spadesuit$, Lizhen Qu$^\spadesuit$\thanks{~~Corresponding author: \texttt{lizhen.qu@monash.edu}}} \\
$^\spadesuit$ Faculty of Information Technology, Monash University, Australia \\
$^\diamondsuit$ School of Information Technology, Monash University Malaysia \\
$^\heartsuit$ School of Business, Monash University Malaysia \\
$^\clubsuit$ Faculty of Law, Monash University, Australia \\
\texttt{\{firstname.lastname\}@monash.edu}
}

\maketitle
\begin{abstract}
More than half of the global population struggles to meet their civil justice needs due to limited legal resources. While Large Language Models (LLMs) have demonstrated impressive reasoning capabilities, significant challenges remain even at the foundational step of legal issue identification.
To investigate LLMs' capabilities in this task, we constructed a dataset from 769 real-world Malaysian Contract Act court cases, using GPT-4o to extract facts and generate candidate legal issues, annotated by senior legal experts, which reveals a critical limitation: while LLMs generate diverse issue candidates, their precision remains inadequate (GPT-4o achieves only 62\%).
To address this gap, we propose \ourname (\uline{Le}gal \uline{P}rofessional-inspired \uline{R}easoning \uline{E}licitation and \uline{C}lassification), a neuro-symbolic framework combining neural generation with structured statistical reasoning. \ourname consists of: (1) a \textbf{neural} component leverages LLMs to transform legal descriptions into question–answer pairs representing diverse analytical factors, and (2) a \textbf{symbolic} component applies sparse linear models over these discrete features, learning explicit algebraic weights that identify the most informative reasoning factors. Unlike end-to-end neural approaches, \ourname achieves interpretability through transparent feature weighting while maintaining data efficiency through correlation-based statistical classification. Experiments show a 30–40\% improvement over advanced LLM baselines, including GPT-4o and Claude, confirming that correlation-based factor–issue analysis offers a more data-efficient solution for relevance decisions. \href{https://github.com/fanyuuwang/LePREC-Reasoning-as-Classification-over-Structured-Factors-for-Assessing-Relevance-of-Legal-Issues}{\faGithub\ }

\end{abstract}

\section{Introduction}
\label{sec:intro}

Access to justice remains a global challenge, with over half of individuals worldwide unable to meet their civil justice needs~\cite{justiceReport}. The shortage of legal expertise underscores the need for automated legal reasoning tools. Within the IRAC framework (Issue, Rule, Application, Conclusion)~\cite{stockmeyer2021legal,kang2023can}, legal issue identification, comprising both generating candidate issues and assessing their relevance, is the crucial first step, determining which legal questions arise from given facts. Given a set of facts, large language models (LLMs) show promise for \textit{generating candidate legal issues} due to their strong language capabilities~\cite{siino2025exploring,bernsohn2024legallens}, however, their assessment of \textit{issue relevance} remains imprecise in real-world contexts~\cite{schroeder2023one,magesh2025hallucination}. 

\begin{figure*}
    \centering
    \includegraphics[width=0.9\textwidth]{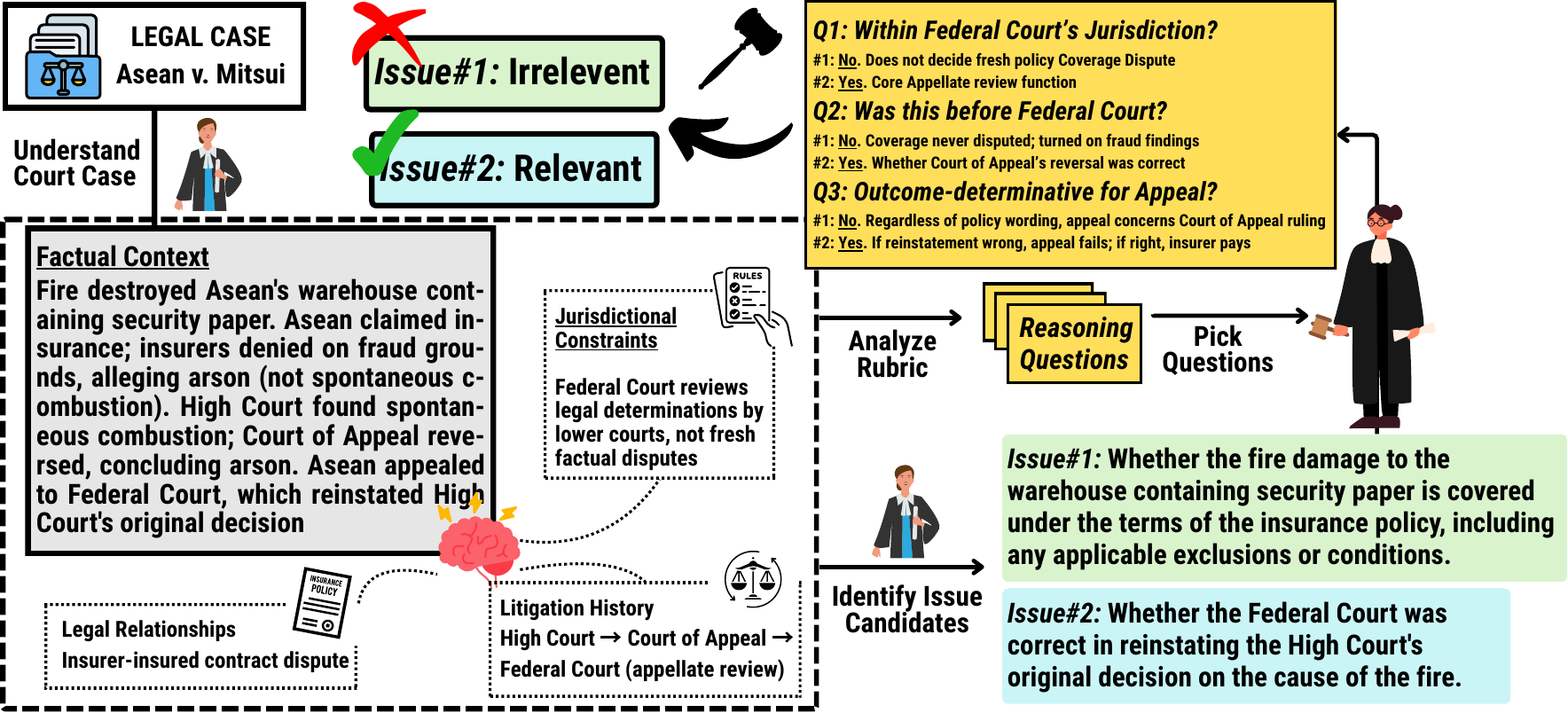}
    \caption{Legal issue relevance assessment requires understanding multiple contextual layers beyond surface-level fact matching. In this appeal, a candidate issue about policy coverage appears factually relevant but is \textbf{irrelevant} because the Federal Court (an appellate body) does not make fresh policy determinations. The truly \textbf{relevant} issue addresses what this specific court is being asked to decide: whether the lower court's ruling was legally correct. Lawyers employ reasoning questions that capture jurisdictional constraints, procedural context, and case-specific factors to distinguish relevant from irrelevant, indicating that a nuanced judgment requires expert legal knowledge.}
    \label{fig:working_example}
\end{figure*}


Assessing the relevance of generated issues requires both realistic legal scenarios and expert evaluation, yet such resources remain scarce. Existing benchmarks are limited to simplified or synthetic settings: ~\citet{guha2023legalbench} avoid full legal case transcripts, while~\citet{kang2024bridging} evaluate zero- and few-shot LLM performance on textbook-derived examples. The lack of real-world court case datasets hinders rigorous evaluation of models’ ability to identify legally relevant issues, leaving a major gap in assessing their practical utility~\cite{magesh2025hallucination}.

To enable rigorous evaluation, we curate a \underline{L}egal \underline{I}ssue dataset on \underline{C}ourt cases, coined \ourdata, comprising 769 Malaysian Contract Act decisions. Herein, GPT-4o is used to extract facts and generate issue candidates, which are then annotated for relevance by senior lawyers, legal academics, and advanced law students. This is the first expert-validated, real-world benchmark for legal issue relevance assessment. Our analysis on this dataset shows that while LLMs generate diverse issue candidates, their precision remains low, simply prompting state-of-the-art LLMs achieves only 62\% precision. The difficulty lies in reasoning beyond surface similarity among facts, as legal professionals assess relevance by considering multiple factors, including the core controversy and legal principles within the correct jurisdictional and procedural context ( Fig.~\ref{fig:working_example}).

Legal professionals approach such assessments through a natural two-stage process: i) they \textit{identify key analytical factors} by brainstorming jurisdictional constraints, procedural context, and case-specific considerations; ii) they \textit{weigh these factors} to reach a final relevance judgment. This decomposition, separating factor extraction from factor-based reasoning, mirrors the neuro-symbolic paradigm of combining neural comprehension with symbolic deliberation, motivating the design of our framework, named \ourname (\uline{Le}gal \uline{P}rofessional-inspired \uline{R}easoning \uline{E}licitation and \uline{C}lassification). It reframes the reasoning-based relevance assessment task as a neuro-symbolic framework combining neural factor extraction with statistical classification over symbolic features. \ourname consists of a two-stage framework: i) the \textbf{neural component} leverages LLMs' language understanding to generate and answer rich reasoning Yes/No questions, transforming unstructured fact-issue descriptions into symbolic features capturing diverse analytical factors, and ii) the \textbf{symbolic component} employs sparse linear models to learn explicit algebraic weights over these discrete features, identifying the most informative reasoning cues for relevance assessment. This formulation transforms the task from evaluating fact–issue relationships to assessing factor–issue relevance, enabling similarity-based reasoning at an abstraction level, where similar factors yield similar relevance judgments. It is data-efficient because the number of model parameters is comparable to the size of the training data. Our contributions are threefold:

\begin{itemize}[nosep]
    \item We present \ourdata, the \textit{first} large annotated dataset for legal issue relevance assessment, comprising 769 real-world court cases drawn from authentic judicial decisions. 
    \item We propose \ourname, a neuro-symbolic approach that mirrors legal professionals' analysis, transforming text-based legal reasoning into interpretable statistical classification over structured features by separating neural comprehension from symbolic deliberation.
    \item We conduct extensive experiments demonstrating that \ourname\ achieves 30-40\% improvement over cutting-edge LLM baselines. Ablation studies confirm the effectiveness of structured features and sparse linear models for relevance assessment. The interpretable statistical analysis of linear models offers deeper insight into how informative QA pairs contribute to the final relevance decisions. 
\end{itemize}
\section{The \ourdata\ Dataset}
\label{sec:dataset}

To enable rigorous evaluation of legal issue relevance assessment in real-world settings, we construct \ourdata, a corpus built from 769 Malaysian Contract Act court cases. Malaysia's Commonwealth legal system shares \citetalias{commonwealth_members} with 50+ jurisdictions, including the UK, Australia, and Singapore~\cite{greenleaf2013building}. Contract law exhibits universal reasoning principles, jurisdictional constraints, procedural posture, and factual relevance that transcend specific legal systems~\cite{kazantsev2022civilizational}. These characteristics suggest strong generalization ability in assessing legal understanding and judgment for the targeted legal analysis approaches. This section describes our dataset construction process, emphasizing the diversity of issue generation and the critical precision challenges revealed through expert annotation (The algorithm is illustrated in Appendix \ref{app:data_algo}).

\paragraph{Dataset Construction Overview.}
We collect 769 court cases from the \textit{\citetalias{clj} (CLJ)}, focusing on \textit{illegality under Section 24 of the Contracts Act Malaysia} and the \textit{formation of contracts}. Starting with 243 Federal and High Court judgments, prioritized for their citation reputation, we expand the dataset by tracing cited cases within each judgment, yielding approximately 20 related cases per primary case. The dataset spans judgments from the 1990s to the present, capturing diverse legal scenarios and judicial writing styles.

We employ \gpto\ to automatically extract facts and legal issues from case PDFs, following best practices in prompt engineering~\cite{wang2024prompt,lin2023unlocking}. This process yields 5,690 issues and 7,397 facts (Prompt templates in Appendix~\ref{app:prompt_extraction}). \rev{We refer to these GPT-extracted issues as \textit{silver truth issues} to distinguish them from the \textit{ground truth labels}, i.e., binary relevance annotations (Relevant / Irrelevant) assigned by senior legal experts and used as the supervision signal for classification. A single datapoint consists of a (fact set, candidate issue) pair as input and a ground truth label as output; at inference time, the model observes only the fact set and candidate issue, and predicts its binary relevance label.} To validate extraction quality, we engage a team of four annotators, including junior lawyers and high-performing law students, who evaluate randomly sampled outputs against original case documents using structured criteria. \textit{65.1\% of outputs achieve ``High Distinction'' ratings and 30.2\% receive ``Pass'' ratings, with facts demonstrating the highest inter-annotator agreement} (Detailed analysis in Appendix~\ref{app:evaluationGuide}).

\paragraph{Incremental Issue Generation for Diversity.}
\label{sec:issue_gen}
While extracted issues from case documents are reliable, legal disputes often admit multiple valid interpretations. A single case may yield different perspectives depending on how issues are framed, making diverse issue generation essential. Moreover, negative (non-applicable) issues play a crucial role in training, as even correctly extracted issues are not uniquely determined by facts.

Inspired by the principle that LLMs perform best with sufficient and necessary information~\cite{feng2023less}, we propose an \emph{incremental generation strategy} to produce diverse issue candidates while avoiding spurious correlations. Given a fact list $\mathbf{X} = \{\mathbf{x}_1, \ldots, \mathbf{x}_m\}$, we generate incrementally:
\begin{enumerate}[nosep]
    \item Generate issues $\mathcal{\hat{Y}}_1$ using only $\mathbf{x}_1$, initializing $\mathcal{\hat{Y}} = \mathcal{\hat{Y}}_1$.
    \item Generate issues $\mathcal{\hat{Y}}_2$ using $[\mathbf{x}_1, \mathbf{x}_2]$, updating $\mathcal{\hat{Y}} = \mathcal{\hat{Y}} \cup \mathcal{\hat{Y}}_2$.
    \item Repeat by incrementally adding facts until all are used: $\mathcal{\hat{Y}} = \bigcup_{i=1}^{m}\hat{\mathcal{Y}}_{i}$.
\end{enumerate}
By varying the ``depth'' of context, this strategy encourages the LLM to attend to progressively richer fact combinations, uncovering nuanced issue candidates that single-pass generation might miss. We craft prompts using \gpto\ and refine them with Claude (see Appendix~\ref{app:incremental_issue_gen_prompt}).

Accordingly, we compare against a \emph{baseline} that feeds all $m$ facts to the LLM in one pass with $m$ times sampling, producing $\hat{\mathcal{Y}}_{m}\subseteq\hat{\mathcal{Y}}=\bigcup_{i=1}^{m}\hat{\mathcal{Y}}_{i}$. We evaluate both \textbf{quality} (whether generated issues semantically cover \rev{silver truth issues}) and \textbf{diversity} (whether candidates differ meaningfully from each other). We employ five metrics: Fréchet BERT Distance (FBD) and BERT Embedding Distance (EMBD)~\cite{alihosseini2019jointly} for quality, and Self-EMBD, Self-BLEU, Distinct-N~\cite{zhu2018texygen, li2015diversity} for diversity. Table~\ref{tab:issue_gen} shows that our incremental method substantially outperforms the baseline, achieving better coverage of \rev{silver truth issues} while generating more diverse candidates (Detailed in Appendix~\ref{app:diversity}).

\begin{table*}[t]
\centering
\caption{Results of Issue Generation Strategies}
\vspace*{-0.5em}
\label{tab:issue_gen}
\small
\begin{tabular}{lcccccc}
\toprule
\multirow{2}{*}{\textbf{Methods}} & \multicolumn{2}{c}{\textbf{Quality}} & \multicolumn{3}{c}{\textbf{Diversity}} \\ 
\cmidrule(lr){2-3} \cmidrule(lr){4-6}
& \textbf{FBD} & \textbf{EMBD} & \textbf{Self-EMBD} & \textbf{Self-BLEU (n=3,4,5)} & \textbf{Distinct-N (n=3,4,5)} \\
\midrule
Baseline & 1311 & 1354 & 211.8 & 11.85/14.52/16.64 & 8250/8559/8604 \\
Ours     & \textbf{1177} & \textbf{1227} & \textbf{225.1} & \textbf{24.90/30.36/34.95} & \textbf{11341/11790/11766} \\
\bottomrule
\end{tabular}
\vspace*{-1em}
\end{table*}

\paragraph{Expert Annotation of Issue Relevance.}
Legal issue relevance assessment is inherently subjective, with experts often reaching divergent conclusions. To establish consistent annotation standards, we convened multiple discussion rounds with three legal scholars, each with over 15 years of courtroom or academic experience, resulting rubric centers on a core principle: \textit{an issue is deemed relevant only if it directly addresses the main dispute or core facts of the case, not merely if it relates to the scenario}, which ensures annotators capture precise factual-legal linkages rather than restating background principles. All the annotation processes are under regulation, with detailed explanation in Appendix~\ref{app:relevance_annotation}. We recruited three annotators with strong backgrounds in Commonwealth law jurisdictions through formal interviews, led by a senior solicitor with seven years of practice. Each annotator independently evaluated issue-fact pair relevance in the test set. Inter-annotator agreement measured by Fleiss' $\kappa$ is 0.659, with pairwise Cohen's $\kappa$ scores of 0.647, 0.746, and 0.584, indicating substantial agreement despite the task's inherent subjectivity~\cite{gwet2008computing, mchugh2012interrater}.

\paragraph{Dataset Statistics and LLM Precision Analysis.}
The \ourdata\ corpus comprises two subsets: a large unlabeled collection (\unlabeled), consisting of extracted issue and fact pairs from court judgments treated as highly relevant by construction, and a rigorously annotated test set (\labeled), \rev{where each (fact set, candidate issue) pair carries an expert ground truth label}. Table~\ref{tab:stat_dataset} summarizes key statistics.

\begin{table}[t]
\centering

\caption{Statistics of Issue-Facts Pairs \ourdata Dataset}
\vspace*{-0.5em}
\small

\label{tab:stat_dataset}
\begin{tabular}{lccccc}
\toprule
\textbf{Type} & \textbf{Total} & \textbf{Gold} & \textbf{Rele.} & \textbf{Irrele.} \\
\midrule
\unlabeled & 3,903 & 752 & --  & --  \\
\labeled   & 1,188  &   213 & 607 & 368 \\
\bottomrule
\end{tabular}
\vspace*{-1em}
\end{table}

Through expert annotation, we reveal a critical limitation in current LLM approaches to issue relevance assessment. While our incremental generation strategy successfully produces diverse issue candidates, the \textit{precision of LLM-generated issues remains inadequate}. When we apply state-of-the-art models, including \gpto, Claude, and others to identify relevant issues from the generated candidates, they achieve only 62.26\% precision on \labeled. This low precision stems from LLMs' inability to distinguish between problems that are merely fact-related and those that truly address the case's core controversy, a nuanced judgment requiring deep legal expertise. Even closely fact-related issues may be irrelevant if they fail to address the dispute's central controversy, as demonstrated in Fig.~\ref{fig:working_example}. Unless noted otherwise, all evaluations in this study are performed on \labeled, while the benefits of leveraging \unlabeled\ are explored in Sec.~\ref{sec:methodology}.
\section{Neuro-Symbolic Framework for Assessing Relevance of Legal Issues}
\label{sec:methodology}

\begin{algorithm}[t]
\caption{\ourname\ Framework}
\label{alg:leprec}
\KwIn{A (fact set, candidate issue) pair $\langle \mathbf{X}, \hat{Y}_j \rangle$}
\KwOut{Binary relevance label $y_j \in \{\text{Relevant}, \text{Irrelevant}\}$}

\tcp{Neural Component: Question Generation}
For each pair $\langle \mathbf{X}_i, \hat{Y}_j \rangle$ in \unlabeled, apply LLM to generate binary reasoning questions $\mathcal{Q}_{i,j} = \{q_1, \ldots, q_h\}$\;
Accumulate into shared question pool $\mathcal{Q} = \bigcup_{i,j} \mathcal{Q}_{i,j}$, where $h = |\mathcal{Q}| = 2{,}486$\;
\textit{Note: $\mathcal{Q}$ is generated from \unlabeled\ and shared across all cases}\;

\tcp{Neural Component: Question Answering}
For each question $q_t \in \mathcal{Q}$, apply generative verifier $G$ to compute answer probability $G_{q_t}(\mathbf{X}, \hat{Y}_j) \in (0, 1)$\;
Collect scores into feature vector $\mathbf{f} = G_{\mathcal{Q}}(\mathbf{X}, \hat{Y}_j) \in \mathbb{R}^h$\;

\tcp{Symbolic Component: Linear Classification}
Predict relevance label via learned linear model:
$\hat{y}_j = \text{sign}(\mathbf{w}^\top \mathbf{f})$\;

\KwRet{$\hat{y}_j \in \{\text{Relevant}, \text{Irrelevant}\}$}
\end{algorithm}

We operationalize legal professionals' factor-based reasoning through a neuro-symbolic framework: neural question generation transforms unstructured legal text into symbolic features, followed by a linear classifier for interpretable relevance assessment over a structured feature space.

\paragraph{Problem Formulation.}
Given a fact set $\mathbf{X}=\{\mathbf{x}_{1},\dots,\mathbf{x}_{m}\}$ and an issue candidate $\hat{\mathcal{Y}}$, our goal is to predict a binary relevance label indicating whether $\hat{\mathcal{Y}}$ is legally germane to $\mathbf{X}$.

\paragraph{Opening Challenges.}
The annotation results in Sec.\ref{sec:dataset} indicate that LLMs struggle with nuanced judgments regarding relevance in identifying legal issues. Our preliminary results in Sec.\ref{sec:results} demonstrate that LLMs not only have difficulty generating relevant issue candidates but also identifying them accurately: Claude and GPT-4o achieve only $\approx$55\% and  $\approx$58\% F1-Score, respectively. These shortfalls underscore a central challenge: \emph{the conventional ``black-box'' strategy of mapping inputs directly to outputs falls short, whereas neuro-symbolic methods that mirror legal professionals' step-wise reasoning hold greater promise.}

\paragraph{Methodology Inspiration.}
By consulting legal experts, we observe their analysis follows a two-stage process with three operations: i) generating comprehensive analytical questions or rubrics (\textit{factor extraction}), ii) selecting a focused subset pertinent to case context (\textit{factor selection}), and iii) making decisions based on selected questions (\textit{factor weighting}), where we provide a case study of \ourname in Appendix~\ref{app:case_study} aligned with Fig.\ref{fig:working_example}.

This naturally embodies neuro-symbolic decomposition: operation (i) requires language understanding to extract structured factors from unstructured text (\textit{neuro}), while operations (ii)-(iii), which select informative factors and learn their weights, constitute symbolic deliberation (\textit{symbolic}) through explicit algebraic operations (L1 regularization for selection, linear weighting for assessment), distinguishing them from neural black-box processing.

\paragraph{\textit{Neural} Generation of Reasoning Questions.}
Legal professionals examine issues by eliciting diverse reasoning questions representing different rubrics and concerns. We operationalize this through LLMs, leveraging their language understanding to generate sets of binary questions on different issue-fact pairs, forming question pool $\mathcal{Q}$:
\begin{enumerate}[nosep]
  \item Apply the LLM to produce several contextualized questions $\mathcal{Q}_{i,j}=\{\mathbf{q}_1,..., \mathbf{q}_h\}$ for pairs of facts $\mathbf{X}_i$ and corresponding legal issue $\mathcal{Y}_j$. Then update the question pool $\mathcal{Q}=\mathcal{Q} \cup \mathcal{Q}_{i,j}$.
  \item Given a question $\mathbf{q}_{t}\in\mathcal{Q}$ and a pair $\langle\mathbf{X}_{i},\mathcal{Y}_{j}\rangle$, generation method $\mathcal{G}$ outputs $\mathcal{G}_{\mathbf{q}_{t}}(\mathbf{X}_{i},\mathcal{Y}_{j})\in(0,1)$. Collectively, these scores form an $h$-dimensional feature vector $\mathcal{G}_{\mathbf{Q}}$ for each data point, where $h=|\mathcal{Q}|$.
\end{enumerate}
We generate questions from \unlabeled (since extracted issues in \labeled are insufficient), resulting in 2,486 questions. For the generation method $\mathcal{G}(\cdot)$, we adopt probability-based Generative Verifier~\cite{zhang2024generative} rather than direct binary LLM responses, as preliminary results indicate direct answers are unreliable (Detailed in Sec.~\ref{sec:results}).

Although only \gpto is used, question generation is a model-agnostic process that can be applied to different LLMs, as subsequent sparse feature selection automatically retains only the most predictive factors with observable weights, thereby significantly reducing the model-specific noise. This neural stage transforms unstructured legal text into structured features, creating a symbolic-like representation where each dimension corresponds to one analytical factor.

\paragraph{Correlation-Aware \textit{Symbolic} Prediction.}
The question generation creates a large pool of diagnostic questions covering a wide range of analytical perspectives. However, using all questions directly presents two challenges. \textbf{\uline{Challenge 1}}, semantically similar questions are expected to collaboratively yield consistent results. However, due to LLM unreliability, similar questions often produce conflicting outcomes, turning potential collaboration into noise. \textbf{\uline{Challenge 2}}, some questions are highly domain-specific, leading to noise when applied to unrelated fact-issue pairs. E.g., the question \textit{``Is the issue central to the insurer's stated reason for denying the claim?''} pertains only to insurance disputes. We cannot simply discard these narrow questions, as doing so would eliminate crucial information when their domain is relevant.

We adopt linear models on symbolic features to demonstrate how explicit algebraic operations on analytical factors address the identified challenges. Linear models exemplify key symbolic properties, explicit weight coefficients and transparent algebraic combination, while offering practical advantages: competitive performance with data efficiency and interpretable analysis of which specific questions contribute to legal reasoning, how their weights reveal relative importance, and how they activate across different case contexts. Through learned coefficients, linear models implement correlation-based feature weighting where the optimization process distributes weights according to marginal contributions. The linear function $\mathbf{w}^\top \mathbf{f}$ learns a weighted combination of structured features
, which addresses \textbf{\uline{Challenge 1}} by automatically downweighting noisy or redundant features through learned coefficients, treating correlated features as collective signals. For \textbf{\uline{Challenge 2}}, standard linear models employ adaptive weighting without removal: retaining all features with non-zero weights ($w_j \neq 0$), learning coefficients that reflect marginal contributions. The linear combination enables implicit contextualization where domain-specific features amplify when relevant context is present and attenuate otherwise. In contrast, L1-regularized variants induce sparsity by setting $w_j = 0$ for selected features, eliminating them globally, creating tension with preserving domain-specific questions needed in specialized contexts. We empirically compare these strategies to evaluate their trade-offs in handling correlated features while maintaining symbolic interpretability.

\section{Experiments}
\label{sec:results}
We systematically evaluate our correlation-aware framework through three research questions: 
\textbf{RQ1}: \emph{How well do state-of-the-art LLMs perform on legal issue relevance classification?} We benchmark cutting-edge LLMs on \ourdata\ to establish baseline performance and reveal fundamental limitations of direct LLM judgment approaches.
\textbf{RQ2}: \emph{How does our correlation-aware linear framework compare against baselines?} We evaluate \ourname\ against alternatives, demonstrating that correlation-aware linear weighting outperforms sophisticated end-to-end LLM approaches.
\textbf{RQ3}: \emph{What is essential in relevance classification?} We investigate whether stable, privileged questions drive legal reasoning through stability analysis of feature selection methods across multiple configurations, and verify our findings through an interview-based analysis with legal practitioners.

\paragraph{Experimental Setup.}
We report the mean and standard deviation of accuracy, macro-F1, macro-precision, and macro-recall under stratified 5-fold cross-validation. We use \labeled\ with an expert-annotated test set and split the remaining data into training and validation sets at a 70:30 ratio. Hyperparameter details are provided in Appendix~\ref{app:hyperparameters}.

\paragraph{RQ1: SOTA Methods on \ourdata.}
We evaluate two types of methods: reward models and LLMs as judges. Detailed settings and prompt templates appear in Appendix~\ref{app:hyperparameters} and \ref{app:prompt_preliminary}.

\subpara{SOTA Baselines}\textbf{Reward Models}: (i) Generative verifier~\cite{zhang2024generative}, simplifying reward tasks to next token prediction on ``yes'' and ``no'' tokens (``Gen''); (ii) Prometheus~\cite{kim2024prometheus}, an open-source evaluation model using absolute rewarding mode (``Prom''); (iii) BERT-based Classifier~\cite{chalkidis2020legal}, a legal-specialized pre-trained LM with binary classifier head (``LBERT''). We initialize models with \citetalias{gptoss20b} (``oss''), \citetalias{microsoftphi4} (``Phi''), and \citetalias{yang2025qwen3} (``Qwen''). \textbf{Large Language Models}: Following LLMs-as-judge~\cite{zheng2023judging}, we evaluate \gpto\ and \citealp{anthropic_link} (``GPT4o'' and ``Claude'') (Prompt in Appendix~\ref{app:prompt_preliminary}). Besides, we also use \citetalias{gptoss20b}, \citetalias{microsoftphi4} and \citetalias{yang2025qwen3} as LLMs as judges.

\begin{table}[!t]
\footnotesize
\centering
\caption{Comparison of Current LLMs on \ourdata (RQ1)}
\vspace*{-0.5em}
\label{tab:preliminary}
\resizebox{\columnwidth}{!}{%
\begin{tabular}{ccccc}
\hline
Methods                & F1 & Acc. & Prec. & Rec. \\ \hline
Claude                 & $54.55_{\pm 1.85}$    & $70.91_{\pm 1.16}$     &  $66.00_{\pm 3.48}$    &  $56.19_{\pm 1.36}$  \\
GPT4o                  & $57.80_{\pm 1.98}$    & $70.91_{\pm 1.16}$     &  $64.46_{\pm 2.52}$    &  $58.07_{\pm 1.56}$  \\
GPT-OSS                & $40.51_{\pm 0.11}$    & $68.10_{\pm 0.30}$     &  $34.37_{\pm 0.01}$    &  $49.33_{\pm 0.23}$  \\
Phi-4                  & $54.03_{\pm 2.65}$    & $58.59_{\pm 2.60}$     &  $54.12_{\pm 2.50}$    &  $54.50_{\pm 2.74}$  \\
Qwen3                  & $55.33_{\pm 2.62}$    & $71.63_{\pm 1.31}$     &  $68.73_{\pm 4.39}$    &  $56.91_{\pm 1.81}$  \\
LBERT                  & $52.31_{\pm 13.4}$    & $41.28_{\pm 7.91}$     &  $52.10_{\pm 5.52}$    &  $50.79_{\pm 2.14}$  \\
Prom                   & $48.63_{\pm 1.74}$    & $62.76_{\pm 1.31}$     &  $50.05_{\pm 2.24}$    &  $44.96_{\pm 1.81}$  \\
Gen$_\text{oss}$       & $45.15_{\pm 2.47}$    & $51.96_{\pm 5.31}$     &  $44.73_{\pm 1.84}$    &  $50.03_{\pm 1.45}$  \\
Gen$_\text{Phi}$       & $54.03_{\pm 2.65}$    & $58.59_{\pm 2.60}$     &  $54.12_{\pm 2.50}$    &  $54.50_{\pm 2.74}$  \\
Gen$_\text{Qwen}$      & $63.70_{\pm 3.15}$    & $68.59_{\pm 3.34}$     &  $63.84_{\pm 3.30}$    &  $63.92_{\pm 3.03}$  \\
\hline
\end{tabular}%
}
\vspace{-1em}
\end{table}

\subpara{RQ1 Results} Table~\ref{tab:preliminary} shows Gen$_\text{Qwen}$ achieves the highest F$_1$ (63.70\%), followed by GPT-4o (57.80\%) and Qwen3 (55.33\%). The generative verifier framework's performance varies substantially with backbone model quality, as Gen$_\text{Qwen}$ outperforms Gen$_\text{Phi}$ (54.03\%) and Gen$_\text{oss}$ (45.15\%). LegalBERT exhibits high variance (F$_1$ = 52.31$_{\pm13.4}$) due to its thirst for training data. Critically, even the best-performing method achieves only 63.70\% F$_1$, indicating they struggle in understanding the boundary of relevance in legal issue identification.
\vspace*{-0.5em}
\find{\textbf{Answer to RQ1:} Current LLMs cannot precisely judge the relevance of issue candidates.}

\paragraph{RQ2: \ourname\ on \ourdata.}
We systematically evaluate alternatives at each key step and present results in Table~\ref{tab:rq2}\footnote{For each method, we run results on three feature generation models and report their best performance. Check Appendix~\ref{app:full_rq2} for all results.}. \textbf{RQ2-1}: \emph{How does reasoning question generation improve relevant issue identification?} We compare classification methods with SOTA LLMs to validate reasoning question generation efficacy. \textbf{RQ2-2}: \emph{How do different selection methods handle legal reasoning questions?} By comparing with RQ2-1, we investigate the L1 regression and LLM-based selection methods to explore whether the feature selection can maintain a smaller size with comparable performance. RQ2 primarily focuses on the research-level performance. We further provide a discussion from the deployment-level in Appendix \ref{app:deployment}.


\begin{table}[t]
  \footnotesize
  \centering
  \caption{The Bests of the Alternatives (RQ2)}  
  \vspace*{-0.5em}
  \label{tab:rq2}
  \resizebox{\columnwidth}{!}{%
    \begin{tabular}{lcccc}
      \hline
      Methods & F1 & Acc. & Prec. & Rec. \\ 
      \hline
      \multicolumn{5}{c}{RQ2-1: Question Generation}\\ 
      \hline

    \text{\text{SVC$_\text{Phi}$}}  
        & $\mathbf{{80.19}}_{\pm\text{2.83}}$ 
        & $\text{82.66}_{\pm\text{2.38}}$ 
        & $\text{79.67}_{\pm\text{2.70}}$ 
        & $\mathbf{{81.01}}_{\pm\text{3.13}}$ \\

    \text{LR$_\text{Phi}$}           
        & $\text{79.70}_{\pm\text{2.93}}$ 
        & $\text{82.49}_{\pm\text{2.41}}$ 
        & $\text{79.58}_{\pm\text{2.89}}$ 
        & $\text{80.05}_{\pm\text{3.23}}$ \\ 

    \text{Ridge$_\text{Phi}$}        
        & $\text{80.10}_{\pm\text{2.86}}$ 
        & $\text{82.91}_{\pm\text{2.41}}$ 
        & $\text{80.06}_{\pm\text{2.89}}$ 
        & $\text{80.28}_{\pm\text{3.05}}$ \\

    \text{KNN$_\text{Qwen}$}          
        & $\text{74.53}_{\pm\text{2.06}}$ 
        & $\text{79.12}_{\pm\text{1.81}}$ 
        & $\text{76.06}_{\pm\text{2.61}}$ 
        & $\text{73.66}_{\pm\text{2.01}}$ \\

    \text{RF$_\text{Qwen}$}          
        & $\text{74.45}_{\pm\text{2.94}}$ 
        & $\text{79.63}_{\pm\text{2.74}}$ 
        & $\text{77.52}_{\pm\text{4.56}}$ 
        & $\text{73.04}_{\pm\text{2.42}}$ \\

    \text{DT$_\text{Qwen}$}           
        & $\text{67.81}_{\pm\text{4.88}}$ 
        & $\text{71.79}_{\pm\text{5.57}}$ 
        & $\text{68.72}_{\pm\text{4.76}}$ 
        & $\text{68.67}_{\pm\text{4.08}}$ \\ 

    \text{NB$_\text{Qwen}$}           
        & $\text{65.53}_{\pm\text{4.02}}$ 
        & $\text{69.61}_{\pm\text{3.86}}$ 
        & $\text{65.49}_{\pm\text{3.86}}$ 
        & $\text{66.24}_{\pm\text{3.93}}$ \\

    \text{SGD$_\text{Qwen}$}          
        & $\text{75.24}_{\pm\text{2.94}}$ 
        & $\text{79.37}_{\pm\text{3.08}}$ 
        & $\text{76.65}_{\pm\text{4.39}}$ 
        & $\text{74.58}_{\pm\text{2.28}}$ \\

    \text{\text{LDA$_\text{Phi}$}}  
        & $\text{79.56}_{\pm\text{4.01}}$ 
        & $\text{\text{83.50}}_{\pm\text{2.69}}$ 
        & $\text{\text{81.77}}_{\pm\text{2.97}}$ 
        & $\text{78.39}_{\pm\text{4.30}}$ \\ 
    \hline

    FFN$_{\text{Qwen}}$  
        & $\text{75.65}_{\pm\text{2.57}}$  
        & $\text{79.29}_{\pm\text{2.43}}$  
        & $\text{76.10}_{\pm\text{2.84}}$  
        & $\text{75.57}_{\pm\text{2.47}}$ \\ 

    CNN$_{\text{Qwen}}$  
        & $\text{43.33}_{\pm\text{3.08}}$  
        & $\text{69.78}_{\pm\text{0.96}}$  
        & $\text{54.78}_{\pm\text{24.83}}$  
        & $\text{51.22}_{\pm\text{1.51}}$ \\

    \text{\text{Tranf.$_{\text{Qwen}}$}}  
        & $\mathbf{\text{75.44}_{\pm\text{1.82}}}$  
        & $\text{80.14}_{\pm\text{2.02}}$  
        & $\text{78.22}_{\pm\text{3.64}}$  
        & $\text{74.39}_{\pm\text{2.10}}$ \\

    LSTM$_{\text{Qwen}}$  
        & $\text{63.48}_{\pm\text{2.94}}$  
        & $\text{65.90}_{\pm\text{3.79}}$  
        & $\text{64.04}_{\pm\text{1.93}}$  
        & $\text{65.81}_{\pm\text{1.84}}$ \\

    \text{\text{ResNet$_{\text{Phi}}$}}  
        & $\mathbf{\text{67.33}_{\pm\text{3.61}}}$  
        & $\text{72.48}_{\pm\text{4.19}}$  
        & $\text{68.84}_{\pm\text{4.89}}$  
        & $\text{67.33}_{\pm\text{3.40}}$ \\
        \hline
      \multicolumn{5}{c}{RQ2-2: Question Selection}\\
        \hline
      L1Reg$_\text{Phi}$  
        & $\text{80.01}_{\pm\text{3.61}}$ 
        & $\mathbf{83.34}_{\pm\text{3.06}}$ 
        & $\mathbf{81.13}_{\pm\text{3.93}}$ 
        & $\text{79.32}_{\pm\text{3.45}}$ \\
              L1SVC$_\text{Phi}$  
        & $\text{77.60}_{\pm\text{2.58}}$ 
        & $\text{80.89}_{\pm\text{1.97}}$ 
        & $\text{77.68}_{\pm\text{2.23}}$ 
        & $\text{77.62}_{\pm\text{2.93}}$ \\ 
      LMSel.$_\text{SVC}$ & $57.78_{\pm1.57}$ & $66.33_{\pm0.82}$ & $58.90_{\pm1.20}$ & $57.64_{\pm1.52}$ \\
      LMSel.$_\text{L}$ & $45.63_{\pm1.79}$ & $50.25_{\pm1.72}$ & $46.27_{\pm1.76}$ & $45.84_{\pm1.92}$ \\ 
      LMSel.$^\text{LR}_\text{P}$                       & $74.86_{\pm2.85}$ & $77.44_{\pm2.71}$ & $74.24_{\pm2.71}$ & $76.32_{\pm2.85}$ \\
      GCI                    & $64.32_{\pm 5.29}$    & $66.50_{\pm 7.18}$     &  $69.43_{\pm 8.96}$    &  $65.52_{\pm 3.53}$  \\
GCI$_\text{Chain}$     & $64.07_{\pm 5.16}$    & $69.20_{\pm 2.38}$     &  $72.55_{\pm 6.11}$    &  $65.32_{\pm 4.10}$  \\
GCI$_\text{Cons}$      & ${67.63}_{\pm 8.09}$ & ${74.19}_{\pm 5.05}$ & ${80.72}_{\pm 5.62}$ & ${68.11}_{\pm 6.37}$ \\
      \hline
    \end{tabular}%
  }
  \vspace*{-1em}
\end{table}

\subpara{Alternative Methods} From RQ1, three local LLMs (\citetalias{gptoss20b}, \citetalias{microsoftphi4}, \citetalias{yang2025qwen3}) achieve close or better performance than commercial LLMs. We adopt them as $\mathcal{G}(\cdot)$ for generating features, yielding three alternatives.
\textbf{RQ2-1 Alternatives.} (1) LLM output types: continuous scores (default) or binary labels (\uline{underlined}). (2) Classifier families: \textit{Embedding-based} (FFN, CNN, LSTM, Transformer, ResNet) and \textit{Traditional ML} (SVM, LR, KNN, RF, GB, Ridge, SGD, DT), representing approaches with different inductive biases.
\textbf{RQ2-2 Alternatives.} (1) \textit{L1-regularized methods} (LR$_\text{L1}$, SVC$_\text{L1}$) perform explicit feature selection through sparsity penalties. (2) \textit{LLM-Select (LMSel.)}~\cite{jeong2024llm}: SOTA feature selection leveraging LLMs with standard binary labels (LMSel.), improved variant using probabilities (LMSel.$_\text{P}$), and LLM-based selective classification (LMSel.$_\text{L}$) (Please find the implementation details in Appendix~\ref{app:hyperparameters}). Moreover, CGI~\cite{liu2021everything} is a neuro-symbolic approach that combines causal discovery algorithms with neural networks for legal charge disambiguation. We evaluate three variants: the standard GCI based on causal discovery; GCI$_{\text{Chain}}$ (CausalChain) leverages recurrent neural networks on extracted causal chains; and GCI$_{\text{Cons}}$ (Bi-LSTM+Att+Cons) constrains neural attention using causal strength estimates.

\subpara{RQ2-1 Results} 
Results validate our framework through clear performance stratification. \textbf{Linear models} (Linear SVC, Standard LR, Ridge: 79.70--80.19\% F1) achieve remarkably consistent performance, demonstrating that simple linear weighting addresses \uline{\textbf{Challenge 1}} with data efficiency and interpretability. Near-identical results across variants validate that linear combination itself, not particular regularization, drives performance. \textbf{Tree/distance methods} (RF, KNN, DT: 67--75\% F1) and \textbf{deep learning} show competitive but slightly lower or more variable performance. Feature generator patterns: Phi-4 pairs well with linear models (all $\approx$80\%), while Qwen excels with tree-based and deep methods, aligning with RQ1 where Gen$_\text{Qwen}$ (63.70\%) outperformed Gen$_\text{Phi}$ (54.03\%). Notably, continuous probability scores uniformly outperform binary variants, confirming probabilistic information is critical.

\subpara{RQ2-2 Results}
Feature selection results reveal the challenge of identifying truly important questions. \textbf{LLM-Select} methods fail dramatically, revealing fundamental misalignment: LLMs prioritize questions based on factual salience rather than empirical predictiveness. This bottleneck from RQ1 propagates through selection, demonstrating LLMs cannot reliably identify which questions lawyers find predictive. \textbf{GCI series}, as a hybrid model, even improves over standard causal inference, but still significantly underperforms L1-based methods. As GCI's strict causal discovery overly restricts the feature space, sacrificing significant recall, this confirms that legal relevance relies on a broader set of correlational signals that strict causal graphs may discard, further validating the efficiency of correlation-based sparse selection.

In contrast, \textbf{L1-based methods} achieve competitive performance through selection, while \textbf{L1SVC$_\text{Phi}$} (77.60\%) drops only 2.5 points. This near-parity initially appears to contradict \textbf{\uline{Challenge 2}}, but closer examination reveals that L1 methods eliminate questions based on correlation, including domain-specific questions that contribute to specialized cases, whereas linear models reduce weights rather than eliminate them. Critically, L1's competitive performance raises an intriguing possibility: if sparsity-based selection identifies predictive subsets without relying on LLM judgment, the selected questions and their learned weights may reveal which aspects of legal reasoning drive relevance classification, motivating a deeper investigation into what linear models learn: which questions receive high weights, how weights relate to legal doctrine, and whether coefficient patterns align with human legal reasoning priorities.
\vspace*{-0.5em}
\find{\textbf{Answer to RQ2:} For relevant legal issue identification, $\bullet$ simple linear models with correlation-aware weighting achieve competitive performance (79--80\% F1) while enabling interpretable analysis; $\bullet$ LLMs cannot reliably identify predictive questions through explicit selection; $\bullet$ adaptive weighting outperforms elimination: retention with context-dependent weights preserves domain-specific reasoning.}


\begin{table*}[t]
\centering
\small
\caption{L1 Method Selection Stability Comparison}
\vspace*{-0.5em}
\label{tab:l1_lr_summary}
\begin{tabular}{lccccccc}
\toprule
\multirow{2}{*}{\textbf{Method}} & \multirow{2}{*}{\textbf{Thres.}} & \textbf{Always} & \textbf{Stable} & \textbf{Moderate} & \textbf{Unstable} & \textbf{Never} & \textbf{F1} \\
& & \textbf{(5/5)} & \textbf{($\ge$4/5)} & \textbf{(3/5)} & \textbf{(1-2/5)} & \textbf{(0/5)} & \textbf{Score} \\
\midrule
\multirow{4}{*}{L1 LR} 
& 0.3 & 13 (0.53\%) & 43 (1.75\%) & 71 (2.88\%) & 519 (21\%) & 1831 (74\%) & 73.60$\pm$2.12 \\
& 0.4 & 5 (0.20\%) & 20 (0.81\%) & 24 (0.97\%) & 326 (13\%) & 2094 (85\%) & 72.24$\pm$2.17 \\
& 0.5 & 2 (0.08\%) & 6 (0.24\%) & 18 (0.73\%) & 193 (7.8\%) & 2247 (91\%) & 72.94$\pm$5.31 \\
& 0.6 & 1 (0.04\%) & 3 (0.12\%) & 6 (0.24\%) & 106 (4.3\%) & 2349 (95\%) & 69.89$\pm$4.64 \\
\midrule
\multirow{4}{*}{L1 SVC}
& 0.3 & 386 (15.7\%) & 766 (31.1\%) & 475 (19.3\%) & 1222 (50\%) & 1 (0.04\%) & 76.94$\pm$1.46 \\
& 0.4 & 158 (6.4\%) & 328 (13.3\%) & 327 (13.3\%) & 1729 (70\%) & 80 (3.2\%) & 74.61$\pm$2.66 \\
& 0.5 & 64 (2.6\%) & 153 (6.2\%) & 179 (7.3\%) & 1516 (62\%) & 616 (25\%) & 74.41$\pm$2.03 \\
& 0.6 & 24 (1.0\%) & 61 (2.5\%) & 99 (4.0\%) & 928 (38\%) & 1376 (56\%) & 75.06$\pm$3.23 \\
\bottomrule
\vspace*{-2em}
\end{tabular}
\end{table*}
\paragraph{RQ3: What is Essential in Relevance Classification.}
L1's competitive aggregate performance in RQ2-2 (L1 LR: 80.01\%, L1 SVC: 77.60\%) despite eliminating features raises a critical question: do sparsity-inducing methods identify a stable core of essential reasoning questions? We investigate this through two complementary analyses: a \textbf{quantitative stability analysis} of feature selection methods, and a \textbf{qualitative practitioner study} that grounds our findings in real legal reasoning. Together, they reveal that the absence of a universal question set is not a limitation of our framework, but a fundamental characteristic of legal issue relevance classification itself: even experienced lawyers do not reason from a fixed checklist, but from a broad, context-sensitive coverage of analytical factors.

\subpara{Quantitative: Extreme Selection Instability}
We conduct stability analysis across multiple threshold configurations using 100-iteration bootstrap subsampling within 5-fold CV to investigate whether L1-based selection reveals consistent question subsets. Results reveal no universal essential question set exists, but validate that multiple question subsets can effectively capture legal reasoning (See details in Appendix \ref{app:l1_lr_stable_features}). 

\textbf{L1 Logistic Regression} exhibits extreme instability: only 0.04--0.53\% of features are consistently selected across all 5 folds (Table~\ref{tab:l1_lr_summary}). Critically, these L1-selected features receive near-zero coefficients in Standard LR (e.g., $f_{1914}$: $\beta$=0.061, $p$=0.9998), proving L1 selects arbitrary representatives from correlated clusters rather than genuinely important questions. Hyperparameter sensitivity causes 8--10$\times$ variation in feature counts across folds, yet each fold's different subset achieves reasonable classification, demonstrating many question combinations work equivalently. \textbf{L1 SVC} shows different behavior: 32$\times$ more consistently selected features than L1 LR. Critically, only 38\% overlap between L1 LR and L1 SVC selections proves different methods identify different ``important'' features, yet both achieve similar reasoning capability. This demonstrates that legal reasoning does not depend on privileged questions but on covering sufficient reasoning dimensions.

Three findings explain why no universal essential questions exist: (1) \textit{Extreme instability}: only 0.04--6.4\% consistently selected across methods/thresholds, proving multiple subsets capture legal reasoning equivalently. (2) \textit{Method disagreement}: 62\% non-overlap between L1 LR and L1 SVC selections, yet both function effectively. (3) \textit{Subset equivalence}: 8--11$\times$ variation in feature counts produces similar reasoning capability, validating massive redundancy in our 2,464 questions. Our contextualized question generation ($\approx$8 per case-issue pair) explains this: questions target specific legal scenarios rather than a universal reasoning bank. While no single golden question set exists, domain-specific questions are essential components of effective subsets. L1 methods arbitrarily eliminate such questions when they appear noisy globally, but these questions contribute valuable signal in specialized contexts. This is why adaptive weighting outperforms elimination-based selection, echoing \textbf{\uline{Challenge 2}}: effective legal reasoning requires comprehensive coverage where domain-specific questions activate conditionally rather than fixed selection of privileged questions.

\subpara{Qualitative: Practitioner Validation}
Practitioner feedback from our human usability study (Appendix~\ref{app:usability}) confirms that the absence of a universal question set reflects how legal reasoning actually works. We picked a court case from \ourdata and invited two legal practitioners with substantial experience in contract law: a lawyer with over 10 years of experience in civil and commercial litigation from China, and a law professor from Australia. These two legal practitioners have extensive practical experience and are from different countries outside Malaysia, thereby demonstrating the generalizability of \ourdata and \ourname in other countries. Despite this, their feedback converges on two key messages that directly align with our quantitative findings. For further details on the study design and full interview transcripts, see Appendix~\ref{app:usability}.

\textit{(i) Lawyers do not reason from a fixed checklist.} Both practitioners confirmed that legal reasoning in practice is context-sensitive and does not follow a prescribed sequence of questions. Lawyer~\#1 noted that not every reasoning question reflects a step they would naturally take in every case. Lawyer~\#2 echoed this, observing that the process is ``much more intuitive than considering a list of particular questions in turn'' and that the relevance of individual questions depends heavily on the nature of the matter. This is consistent with our quantitative finding that no stable universal subset exists, and points to why legal issue relevance classification is fundamentally difficult: relevance judgment requires assembling a context-sensitive combination of factors that varies across cases.

\textit{(ii) Comprehensive coverage is the value, not prescription.} Both practitioners independently identified the same utility in the question list: not as a checklist to follow step by step, but as a resource that ensures comprehensive coverage of considerations that might otherwise be overlooked. Lawyer~\#1 observed that questions they would not spontaneously consider still helped surface case-specific factors, while Lawyer~\#2 noted that the factors provided ``relatively comprehensive coverage'' of key considerations. This directly validates our design choice of retaining a full question pool with adaptive weighting rather than imposing L1 elimination, which discards precisely the domain-specific questions that contribute conditionally.

\vspace*{-0.5em}
\find{\textbf{Answer to RQ3:} The stability and usability results reveal that - What is essential is not a fixed set of privileged reasoning questions, but comprehensive coverage with adaptive weighting. No universal essential question set exists, and this reflects how legal reasoning actually works: lawyers do not reason from a fixed checklist but assemble context-sensitive combinations of factors, which is also what makes relevance classification fundamentally difficult.
}
\section{Related Work}
\paragraph{Feature Selection and Classification.} Feature selection and classification methods include filter, wrapper, and embedded techniques~\cite{Gramegnaetal2022,Alsolamietal2022}, with recent advances using deep learning, reinforcement learning, and evolutionary approaches~\cite{Jiaetal2024,Nguyenetal2024}. LLM-Lasso uses large language models to guide feature selection and enhance robustness by integrating domain knowledge into Lasso regression~\cite{zhang2024llmlasso}. Focus Instruction Tuning enables dynamic control over feature reliance in LLMs through natural-language prompts, improving robustness and fairness~\cite{lamb2024focus}. While traditional classifiers like SVMs and ensemble methods remain widely used~\cite{Bulut2022,Majdoubietal2023}, optimal selection still depends on empirical evaluation tailored to the dataset~\cite{LoyolaFuentesetal2022,Montananaetal2024}.

\paragraph{LLMs in the Legal Domain.} Applying LLMs to legal tasks is challenging due to the complexity of legal knowledge. Studies indicate that current models often capture only surface-level concepts \citep{avelka2023ExplainingLC}, miss crucial legal rule details \citep{Yuan2024CanLL}, and struggle to identify important legal factors \citep{Gray2024UsingLT}. These findings underscore the need for further development before LLMs can function autonomously in legal contexts.
\section{Conclusion}
\label{sec:conclusion}

We evaluate relevance assessment of legal issues by introducing \ourdata, the first expert-annotated dataset from 769 Malaysian Contract Act cases. 
To address the low precision problem of LLMs and the scarcity of training data, we propose \ourname, a legal reasoning-inspired neuro-symbolic approach that transforms text-based reasoning into statistical classification over structured factors, followed by relevance prediction using a sparse linear model, which also supports statistical analysis between factors and issues. 
As a result, \ourname achieves 30-40\% improvement over end-to-end LLM approaches, with linear models substantially outperforming both deep learning methods and direct LLM judgments. Our stability and usability analysis reveals the conclusion from quantitative and qualitative perspectives that no universal essential question set exists in the practical legal reasoning process. Instead, effective performance emerges from comprehensive coverage with adaptive weighting. Our work lays the foundation for theoretically grounded, robust, and interpretable legal AI systems. 
\section{Limitations}

The dataset comprises 769 cases with expert annotations, which naturally reflects the subjective understanding involved in legal issue relevance assessment. Different legal experts may reasonably interpret issue relevance differently based on their experience and perspective. 

Our study focuses on Malaysian Contract Act cases from the Commonwealth legal system. While contract law represents a fundamental legal institution with universal reasoning principles (jurisdictional constraints, procedural context, factual centrality), and \ourname's methodology is jurisdiction-agnostic (learning correlation patterns rather than encoding specific doctrines), empirical validation on additional jurisdictions would further establish cross-system robustness. The Commonwealth legal heritage shared across 50+ nations suggests strong transferability, but civil law systems may require investigation of whether reasoning pattern differences affect framework performance. 

Our dataset is constructed from published judicial opinions, which present facts and issues as refined by judges; testing on party submissions (pleadings, briefs) would be valuable as they are more fact-rich and mirror real-world practice, though such materials are typically not publicly available and would require partnerships with law firms and extensive anonymization procedures. 

Our framework relies on LLM-generated reasoning questions, and while our contextualized generation strategy produces sufficient coverage, exploring alternative question elicitation approaches could provide additional insights. We employ linear models for interpretability and efficiency, which assume linear combinations can capture relevance patterns. The learned weights across our comprehensive question set offer coefficient-level interpretability, though extracting high-level insights from detailed weight distributions requires careful analysis. 

The annotation process involves senior lawyers and legal academics, reflecting the typical resource investment in building specialized legal AI datasets. While \ourname demonstrates human expert-level performance, deployment in real-world legal practice would require additional validation to ensure the system does not introduce biases or disadvantage vulnerable populations.

\section{Ethics Statement}

We acknowledge and adhere to the ACL Code of Ethics throughout our research. All case data are from publicly available court records (Current Law Journal database) and have been carefully handled, with names, unique identifiers, and any potentially offensive language screened, anonymized, or redacted to protect privacy and avoid harmful content. The data collection protocol was approved by the Monash University Human Research Ethics Committee (MUHREC). Annotators were recruited via open advertisement targeting legal professionals with Commonwealth law backgrounds, with formal interviews conducted to ensure qualifications. All annotators provided informed consent after receiving detailed explanations of the annotation task, data usage, and their rights, and were compensated at rates meeting or exceeding local government-mandated standards, appropriate for their expertise level and jurisdiction. Payment terms, training procedures, and the right to withdraw were communicated clearly before participation. We employed AI assistants (GPT-4o: \texttt{gpt-4o-2024-05-13} and Claude 3.5 Sonnet: \texttt{claude-3-5-sonnet-20241022}) for fact and issue extraction, prompt refinement, and incremental generation, documenting all prompts, model versions, sampling parameters, and API settings in corresponding sections, while stressing that final outputs were reviewed and validated by human experts. To mitigate LLM-related risks (hallucination, bias, sensitive content), we applied targeted prompt engineering with few-shot exemplars, maintained human-in-the-loop review, and transparently documented usage. We acknowledge the environmental impact and encourage the community to balance performance gains with sustainability considerations.

\bibliography{custom}
\bibliographystyle{acl_natbib}

\clearpage
\newpage

\appendix
\section{Appendix}

\subsection{Prompt Template of Facts and Issue Extraction}
\label{app:prompt_extraction}
We use the following prompt for fact extraction.
\begin{lstlisting}[style=prompt, caption={Prompt for Fact and Held Extraction}]
You are a legal expert tasked with analyzing a court case.
Your goal is to extract the case name, summarize key legally 
significant facts, and explain the court's final decision (held).

Instructions:
1. **Case Name**: Extract the full official case name. 
   Example: Smith v. Jones [2020] 2 MLJ 35.
2. **Facts**: Identify the facts directly related to the legal 
   issues. Focus on those that establish the dispute, actions, 
   and agreements.
3. **Held (Conclusion)**: Provide the court's final decision, 
   including penalties, remedies, or significant conclusions.

Output Format:
{
    "case_name": "Extracted case name",
    "facts": [
        "Fact 1...",
        "Fact 2..."
    ],
    "held": "Holding or judgment of the court."
}

Case Text:
{case_text}
\end{lstlisting}

The prompt below is used for issue extraction.
\begin{lstlisting}[style=prompt, caption={Prompt for Issue Identification and Application}]
You are a legal expert analyzing a court case.
Your goal is to identify legal issues, apply relevant rules 
to the facts, and provide legal conclusions.

Instructions:
1. Identify each legal issue in the case by framing a question 
   starting with "Whether...".
2. For each issue, apply the relevant rules to the facts using 
   an "if...then" structure.
3. Provide a clear answer (Yes/No or another legal conclusion) 
   for each issue, based on legal reasoning.
4. Multiple applications may be required if more than one rule 
   applies or if multi-step reasoning is necessary.

Output Format:
{
    "issues": [
        {
            "issue": "Whether issue 1...",
            "application": [
                "If [specific fact]... then [application of legal rule]...",
                "If [specific fact]... then [application of another legal rule]..."
            ],
            "answer": "Yes/No or detailed legal conclusion for issue 1..."
        },
        {
            "issue": "Whether issue 2...",
            "application": [
                "If [specific fact]... then [application of legal rule]..."
            ],
            "answer": "Yes/No or detailed legal conclusion for issue 2..."
        }
    ]
}

Example:
- Issue: "Whether the contract is enforceable under Section 24 of the Contracts Act."
- Application:
    - "If the contract is based on illegal consideration, then under Section 24, the contract is void."
    - "If no illegal consideration exists, then under the same section, the contract remains valid."
- Answer: "No, the contract is void due to illegal consideration."

Facts:
{facts}

Rules:
{rules}

Original Case Text:
{case_text}
\end{lstlisting}

\subsection{Dataset Quality Details}
CLJ is a leading Malaysian legal publication providing case law reports, legal commentaries, and statutory updates, serving as a key reference for legal practitioners and researchers. Using predefined filtering criteria. We prioritize Federal and High Court judgments due to their higher citation reputation. Each case was sourced in its original PDF format, preserving the judicial text as delivered.

Starting with an initial set of 243 cases, we expand the dataset by tracing cited cases within each judgment. This citation-based expansion yields approximately 20 related cases per primary case, ultimately increasing the dataset to 769 cases. Spanning judgments from the 1990s to the present, the dataset encapsulates a diverse range of legal scenarios and judicial writing styles.

\begin{figure}
    \centering
    \includegraphics[width=0.9\linewidth]{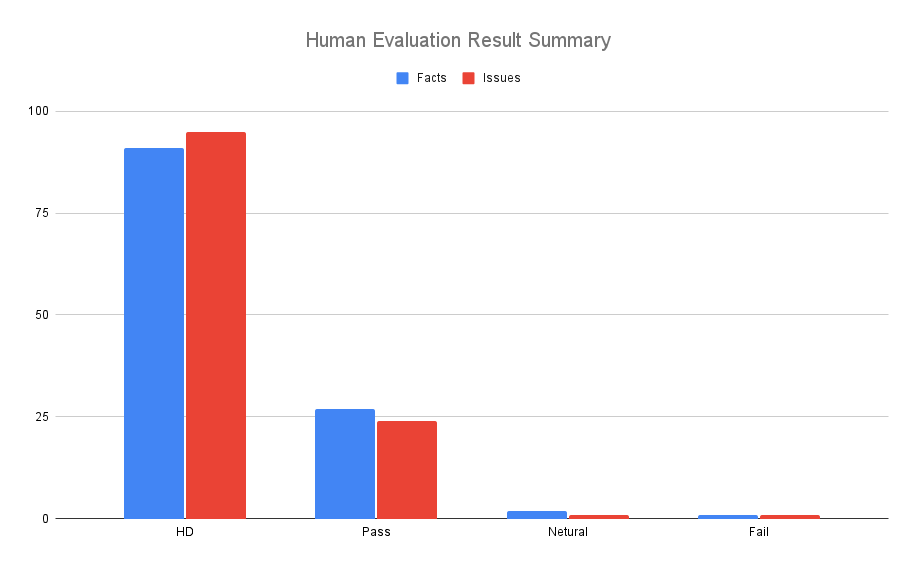}
    \caption{Evaluation results on fact and issue extraction.}
    \label{fig:humanEvlauaitonSummary}
\end{figure} 

\label{app:evaluationGuide}

Legal cases are lengthy, and it is expensive to manually extract facts and issues from those cases. Therefore, we reduce human effort by applying \gpto to extract facts and issues, and by annotating a set of generated issues manually by law students as the ground-truth (see prompt template in Appendix~\ref{app:prompt_extraction}). Herein, the automatically extracted issues are referred to as silver ground-truth. For both fact and issue extraction, we follow the best practice on prompting~\cite{wang2024prompt} and use the styles recommended in~\cite{lin2023unlocking} for prompt design. Specifically, we apply GPT4o with those prompts to extract legally significant facts and legal issues from the PDF files of collected court cases. In the end, we got 5,690 issues and 7,397 facts. \subpara{Data Quality} To check the quality of extracted content, we engage a team of four annotators, including junior lawyers and law students with strong academic records (B+ or higher in relevant legal subjects). Annotators evaluate outputs from randomly sampled cases by comparing them against original content, validating key elements manually. Using predefined criteria, they assign ratings ranging from "High Distinction (HD)", for highly accurate and detailed outputs, to "Fail", for outputs with significant omissions or irrelevance. The detailed guidelines are provided in Appendix~\ref{evaluationGuide}. Structured ratings and detailed comments are provided for each element to assess. As illustrated by Fig.~\ref{fig:humanEvlauaitonSummary}, 65.1\% of the model outputs are rated as HD and 30.2\% as Pass, with facts achieving the highest annotator agreement. Only a small fraction of the facts and issues are categorized as Fail. While HD-rated outputs are ideal, Pass-rated outputs also hold significant value for tasks requiring basic reasoning or tolerating minor inaccuracies. Common errors in Pass-rated outputs include incomplete or poorly sequenced facts, insufficiently framed or misaligned issues. However, even within these outputs, relevant and accurate content often remains, which can be highly beneficial for model training processes. This makes Pass-rated outputs a valuable resource for enhancing dataset diversity and providing foundational reasoning.

\subsection{Human Annotation}
\label{app:relevance_annotation}
\paragraph{Explanation}
Annotator Recruitment and Payment. We recruited annotators through open advertisements and formal calls within our university community, ensuring a diverse pool of participants. Prior to commencing annotation, we obtained approval from our institutional ethics board. All annotators were compensated at or above local government–mandated rates, with payment schedules and amounts clearly communicated in advance. We also provided training sessions, ongoing support, and the opportunity to withdraw at any time without penalty.

\paragraph{Relevance Checking.}
Your task is, for each issue, to decide whether it is “Relevant” or “Irrelevant” to the scenario.

\begin{itemize}
  \item “\textbf{Relevant}”: The issue is not only related to the scenario, but is directly tied to the main dispute or core facts of the case. It goes beyond merely stating a basic legal principle or background fact.
  \item “\textbf{Irrelevant}”: The issue is either unrelated to the scenario or is only a fundamental/basic statement that does not bear on the case’s primary controversy.
\end{itemize}

\begin{table*}[ht]
\caption{Example of the Annotation Answer Sheet. We will both give the extracted facts and ground truth legal issues to help the annotators better capture the core dispute of the case. Then, for each generated legal issue, the annotators need to select ``Relevant'' or ``Irrelevant'' and write several justifications for their decision.}
\resizebox{\textwidth}{!}{%

\renewcommand{\arraystretch}{1.2}
\begin{tabular}{p{3cm} p{3cm} p{12cm}}
\toprule
\textbf{Note} & \textbf{Relevance} & \textbf{Legal Issues} \\
\midrule
\multicolumn{3}{l}{\begin{minipage}{\textwidth}
\textbf{Asean Security Paper Mills Sdn Bhd v. Mitsui Sumitomo Insurance (Malaysia) Bhd [2008] 6 CLJ}
\end{minipage}} \\
\midrule
\multicolumn{3}{l}{\begin{minipage}{\textwidth}
\textbf{Facts:} \\
1. Asean Security Paper Mills Sdn Bhd (the respondent/appellant) made a claim on an insurance policy after a fire destroyed their warehouse containing security paper. \\
2. The insurance companies, including Mitsui Sumitomo Insurance (the applicant/respondent), denied the claim on grounds of fraud, alleging that the fire was caused by arson rather than spontaneous combustion. \\
3. The High Court found that the fire was due to spontaneous combustion; this decision was overturned by the Court of Appeal, which concluded that the fire was an act of arson. \\
4. The respondent/appellant appealed to the Federal Court, leading to the reinstatement of the High Court's original decision. \\[0.5em]
\textbf{Ground Truth Legal Issues:} \\
- Whether the Federal Court should exercise its review jurisdiction under Rule 137 of the Rules of the Federal Court 1995 to prevent injustice in this case. \\
- Whether the findings of fact made by the High Court and reinstated by the Federal Court can be subject to review. \\
- Whether the application for review by Mitsui Sumitomo Insurance was an appropriate use of inherent jurisdiction.
\end{minipage}} \\
\midrule
& \texttt{Irrelevant} & Whether the fire damage to the warehouse containing security paper is covered under the terms of the insurance policy, including any applicable exclusions or conditions. \\
& \texttt{Irrelevant} & Whether Asean Security Paper Mills Sdn Bhd complied with all obligations under the insurance contract when making the claim, such as timely notification and preservation of evidence. \\
\bottomrule
\end{tabular}
}
\end{table*}

\subsection{Dataset Curation Algorithm}
\label{app:data_algo}

See the algorithm list in Algo.\ref{alg:dataset}.

\begin{algorithm}[t]
\caption{Dataset Curation Pipeline}
\label{alg:dataset}
\KwIn{Raw court case PDFs from CLJ}
\KwOut{Labeled set \labeled\ and unlabeled set \unlabeled}

\tcp{Step 1: Fact and Issue Extraction}
Apply \gpto\ to extract facts $\mathbf{X} = \{\mathbf{x}_1, \ldots, \mathbf{x}_m\}$ and silver truth issues $\hat{\mathcal{Y}}$ from each case PDF\;

\tcp{Step 2: Incremental Issue Generation}
\For{$i = 1$ \KwTo $m$}{
    Generate additional issue candidates $\hat{\mathcal{Y}}_i$ using facts $[\mathbf{x}_1, \ldots, \mathbf{x}_i]$\;
    Update $\hat{\mathcal{Y}} = \hat{\mathcal{Y}} \cup \hat{\mathcal{Y}}_i$\;
}

\tcp{Step 3: Expert Annotation}
For each (fact set, candidate issue) pair $\langle \mathbf{X}, \hat{Y}_j \rangle$, collect binary relevance label $y_j \in \{\text{Relevant}, \text{Irrelevant}\}$ from senior legal experts\;

Pairs with expert labels form \labeled; GPT-extracted silver truth pairs form \unlabeled\;
\end{algorithm}

\subsection{Case Analysis and Usability Study}
\label{app:interpretability}

To validate \ourname's interpretability from both a computational and a practitioner perspective, we present two complementary analyses based on a real court case. First, we present a case study and illustrate the reasoning process from \ourname and legal practitioners. Second, we conduct a human usability study with legal practitioners from two distinct legal systems to assess whether the procedure of \ourname is useful across jurisdictions. We present the court case following the structure in \ourdata, as:

\textbf{Court Case Name: Asean Security Paper Mills Sdn Bhd v. Mitsui Sumitomo Insurance [2008] 6 CLJ.} 

\textbf{Fact List:}
\begin{itemize}
    \item The appellant, Encony Development Sdn Bhd, executed a sale and purchase agreement (SPA) with the respondents on 2 September 2010 for a condominium unit, whereby the respondents paid an initial deposit and subsequently the balance ten percent deposit. 

    \item The statutory SPA included clauses that made timely payment of installments essential, allowing the appellant to terminate the agreement for non-payment.

    \item The respondents failed to make progress payments after receiving requests from the appellant and were subsequently issued a notice of default on 12 November 2010, followed by a lawful termination of the SPA on 10 December 2010.

    \item The respondents challenged the termination in the High Court, arguing that there were binding representations made prior to the execution of the SPA which created a collateral contract.
\end{itemize}

\subsubsection{Case Study: Alignment with Expert Legal Reasoning}
\label{app:case_study}
We present a detailed walkthrough of \ourname's factor-based analysis in a Federal Court jurisdiction case, comparing it with annotators' answers to different reasoning questions from \ourdata construction. Note that the annotators didn't review all the reasoning questions, and we use this case study to illustrate the alignment of \ourname and practical legal practitioners' reasoning process. The investigated issues are:
\begin{itemize}
    \item (\textbf{ground truth: Irrelevant}) is: \textit{``Whether the fire damage to the warehouse containing security paper is covered under the terms of the insurance policy, including any applicable exclusions or conditions.''}
\end{itemize}

\paragraph{Expert Reasoning.}
Legal professionals naturally decompose this judgment into a set of reasoning questions:

\begin{itemize}[nosep]
    \item \textbf{Q1} \textit{Is this issue within the Federal Court's jurisdictional scope?} $\rightarrow$ \textbf{No.} The Federal Court hears appeals on questions of law, not fresh coverage disputes.
    \item \textbf{Q2} \textit{Was this specific dispute actually before this court?} $\rightarrow$ \textbf{No.} The dispute concerns whether the Court of Appeal correctly reversed the High Court's factual finding, not policy coverage.
    \item \textbf{Q3} \textit{Does this issue involve fresh factual determinations?} $\rightarrow$ \textbf{Yes.} Interpreting policy terms would require fresh factual analysis the Federal Court does not conduct.
    \item \textbf{Q4} \textit{Does this issue relate to the established facts?} $\rightarrow$ \textbf{Yes.} The issue connects to fire damage facts, providing factual context but not legal relevance.
    \item \textbf{Q5} \textit{Is this a derivative or secondary issue?} $\rightarrow$ \textbf{Yes.} Policy coverage only becomes relevant after determining the cause of fire, which is the primary appellate question.
\end{itemize}

\noindent\textbf{Expert conclusion:} ``This issue is irrelevant because it falls outside the Federal Court's jurisdictional mandate and does not address the core legal question under appeal. While factually connected, it is a derivative concern not before the court.''

\paragraph{\ourname\ Analysis.}
Table~\ref{tab:case_study_features} shows the reasoning questions activated by \ourname's learned model for this issue-fact pair. The linear combination produces a strong negative score ($-0.417$), correctly predicting \textbf{Irrelevant}.

\begin{table}[h]
\centering
\caption{\ourname's active reasoning questions for the insurance coverage issue.}
\small
\resizebox{\columnwidth}{!}{%
\begin{tabular}{lrp{5cm}c}
\toprule
\textbf{Feature} & \textbf{Weight} & \textbf{Reasoning Question} & \textbf{Answer} \\
\midrule
$f_{776}$  & $-0.211$ & Is this issue within Federal Court's jurisdictional scope? & No \\
$f_{1534}$ & $-0.080$ & Was this dispute actually before this court? & No \\
$f_{751}$  & $-0.080$ & Does the issue involve fresh factual determinations? & Yes \\
$f_{440}$  & $+0.122$ & Does the issue relate to established facts? & Yes \\
$f_{2384}$ & $-0.065$ & Is this a derivative/secondary issue? & Yes \\
$f_{1820}$ & $-0.058$ & Does this require interpretation of contractual terms? & Yes \\
$f_{892}$  & $-0.045$ & Would resolving this require examining insurance industry standards? & Yes \\
$f_{1247}$ & $+0.039$ & Does the issue reference specific statutory provisions? & No \\
$f_{563}$  & $-0.042$ & Is this about quantum or extent of damages? & No \\
\bottomrule
\end{tabular}
}
\label{tab:case_study_features}
\end{table}

\begin{align*}
\text{Score} &= (-0.211) + (-0.080) + (-0.080) \\
             &\quad + (+0.122) + (-0.065) + (-0.058) \\
             &\quad + (-0.045) + (+0.039) + (-0.042) + \ldots \\
             &= -0.417 \text{ (strong negative)}
\end{align*}
$\Rightarrow$ \textbf{Prediction: Irrelevant} \checkmark \textit{Correct}

\paragraph{Alignment with Expert Reasoning.}
Three observations confirm alignment between \ourname's learned weights and expert legal reasoning. \textit{(i) Jurisdictional reasoning dominates}: The highest-weighted feature ($f_{776}$, $-0.211$) directly corresponds to the expert's primary consideration, whether the issue falls within the Federal Court's jurisdictional scope. \textit{(ii) Derivative reasoning is captured}: The negative weight on $f_{2384}$ (derivative issue) mirrors the expert's conclusion that policy coverage is secondary to the cause-of-fire question. \textit{(iii) Factual connection is appropriately discounted}: The positive weight on $f_{440}$ (relates to established facts) reflects that the issue is factually connected, but this positive signal is outweighed by the jurisdictional and procedural negatives, consistent with expert reasoning that factual relatedness alone does not establish relevance.

\subsubsection{Human Usability Study}
\label{app:usability}
We further evaluate whether \ourname's reasoning questions are useful to legal practitioners in practice, and whether this utility extends beyond the Malaysian legal context.

\paragraph{Investigated Legal Issue Candidates.}
\begin{enumerate}[nosep]
    \item Whether the alleged pre-SPA representations created a binding collateral contract that could override the terms of the statutory SPA.
    \item Whether the clauses in the statutory SPA making timely payment of instalments essential were valid and enforceable.
    \item Whether the respondents breached a fundamental term of the SPA by failing to make required progress payments.
    \item Whether the notice of default issued on 12 November 2010 was sufficient and in compliance with the terms of the SPA.
    \item Whether the appellant's termination of the SPA was lawful given the respondents' failure to make progress payments.
\end{enumerate}

\paragraph{Reasoning Questions.}
\ourname\ selected 50 reasoning questions from the learned linear model: the top 25 positively weighted questions (indicative of relevance) and the top 25 negatively weighted questions (indicative of irrelevance), listed in Table~\ref{tab:reasoning_questions}. Positively weighted questions target core dispute and outcome-determinativeness, while negatively weighted questions flag peripheral or domain-mismatched considerations, mirroring the interpretability mechanism illustrated in the case study above.

\paragraph{Participants and Procedure.}
We invited two legal practitioners with substantial experience in contract law: a lawyer with over 10 years of experience in civil and commercial litigation from China (Lawyer~\#1), and a law professor from Australia (Lawyer~\#2). Both practitioners are from countries outside Malaysia, thereby providing evidence for the generalisability of \ourdata\ and \ourname\ beyond the Malaysian legal context. Due to time constraints, participants reviewed the reasoning questions holistically across the five issue candidates based on the given facts of the court case and provided feedback via a structured follow-up interview on three questions:
\begin{itemize}
    \item \textit{\textbf{Reasoning Alignment} - To what extent does each factor reflect a reasoning step you would naturally consider when judging whether a legal issue is relevant to the case facts?}.
    \item \textit{\textbf{Discriminative Validity} - Given the facts and different issues, does the factor's answer difference between the different issues align with your own judgment about why one issue is relevant and the other is not?}
    \item \textit{\textbf{Practical Utility} - Do you find these factors useful for you as a practitioner in the legal domain?}
\end{itemize}

\paragraph{Practitioner \#1 (China).}
\subpara{Reasoning Alignment (Q1)}
The practitioner rated overall alignment as high, noting that reasoning around the core dispute was ``almost entirely consistent'' with their own natural reasoning. Coding the interview transcript reveals three key themes. \textit{(i) Core dispute alignment}: Questions targeting the main point of contention (Q3), the rights and obligations of the parties (Q13), and the validity of specific contractual clauses (Q45, Q50) were identified as directly reflective of natural legal reasoning, paralleling the jurisdictional and derivative reasoning identified in the case study above. \textit{(ii) Domain mismatch}: Q33 (\textit{``Is the issue related to a specific professional standard?''}) was flagged as misplaced for a contractual dispute, as the analysis turns on legal provisions and contractual obligations rather than professional conduct norms, consistent with its negative weight in the model. \textit{(iii) Complementary coverage}: The practitioner noted that questions they would not spontaneously consider did not detract from the list; rather, they ensured comprehensive coverage of case-specific considerations. The practitioner further observed that a well-designed reasoning question should be both general enough to apply across cases of the same type and specific enough to accommodate individual circumstances, a balance they found the list maintained well.

\subpara{Discriminative Validity (Q2)}
The practitioner rated discriminative validity as ``Yes,'' confirming that the answer patterns of the reasoning questions were consistent with their own relevance judgments, describing the discrimination as ``professionally competent.'' A concrete example was provided: for Issue~4 (validity of the notice of default), any analysis necessarily involves interpreting specific contractual terms, which aligned precisely with the answer patterns of Q45 and Q50. No reasoning question was found to produce a distinction contrary to the practitioner's judgment.

\subpara{Practical Utility (Q3)}
The practitioner found the reasoning questions practically useful, noting that they ``address real, concrete considerations rather than abstract principles'' and help practitioners ``think more comprehensively and identify the core dispute more directly.'' Q3 was highlighted as particularly effective for Issue~2 (validity of the installment payment clause), focusing attention precisely on the question the court needed to resolve. The practitioner further noted that the list serves a dual function: general questions confirm reasoning the practitioner would already undertake, while case-specific questions serve as useful prompts, reminding practitioners to consider aspects they might not have thought to examine independently.

\paragraph{Practitioner \#2 (Australia).}
\subpara{Reasoning Alignment (Q1)}
The practitioner rated alignment as 2 out of 5 (``to some extent''), noting that in practice they ``would not normally ask all of these'' and would not have a fixed list of questions for any given consideration. The practitioner observed that the actual reasoning process is ``much more intuitive than considering a list of particular questions in turn,'' and that which factors are relevant depends on the nature of the matter. Importantly, however, the practitioner acknowledged that ``the factors that were listed were relatively comprehensive in terms of coverage,'' which is precisely the design intent of \ourname. The lower rating, therefore, reflects the prescriptive format rather than a rejection of the underlying reasoning dimensions.

\subpara{Discriminative Validity (Q2)}
The practitioner rated discriminative validity as ``Partially,'' noting that with such a brief set of facts, it is difficult to identify the appropriate level of relevance without knowing the full procedural context. Specifically, the practitioner observed that it was not immediately apparent why the matter went to the High Court, and that whether this involved first instance or appellate proceedings would materially affect the relevance of several issues. This reflects the same challenge our framework addresses: relevance judgment is inherently context-sensitive, and the difficulty of assessing it from facts alone is a key motivation for structured reasoning question coverage.

\subpara{Practical Utility (Q3)}
The practitioner found the questions useful as a ``catch all'' list that provides ``useful general coverage of many if not all of the key questions,'' echoing Lawyer~\#1's observation about comprehensive coverage. However, the practitioner noted they are not useful as a prescriptive checklist, as many questions would have no specific relevance to a given issue. The practitioner further observed that the list may be of value to junior practitioners learning what to consider, while cautioning that going through a large number of irrelevant questions risks losing concentration. This aligns with our quantitative finding that domain-specific questions contribute conditionally rather than universally, reinforcing the case for adaptive weighting over fixed selection.

\paragraph{Discussion.}
Together, the case study and usability study provide complementary evidence for \ourname's interpretability. The case study demonstrates that the learned weights reproduce expert legal reasoning at the feature level: jurisdictional and procedural questions dominate, factual relatedness is appropriately discounted, and derivative issues are correctly penalized. The usability study extends this to practitioner utility beyond the Malaysian legal context. Despite differences in rating and background, both practitioners independently converged on the same core observation: the value of \ourname's reasoning questions lies in comprehensive coverage rather than fixed prescription, and legal reasoning in practice does not follow a universal checklist. This convergence validates both our framework design and the broader claim that contract law reasoning exhibits universal principles that generalize across jurisdictions (cf.\ RQ3, Sec .~\ref {sec:results}).

\subsection{Discussion on Deployment Readiness}
\label{app:deployment}

While \ourname demonstrates substantial improvements over state-of-the-art LLM baselines, achieving 79.70-80.19\% F1 compared to 55-62\% for GPT-4o and Claude, we acknowledge that the current system represents research-stage performance rather than a production-ready deployment. To properly evaluate this performance gap, we must first contextualize \ourname's results against realistic human expert baselines. Our inter-annotator agreement analysis reveals that three senior legal scholars with over 15 years of experience each achieved Fleiss' $\kappa=0.659$ (``Substantial Agreement''), with pairwise Cohen's $\kappa$ scores ranging from 0.584 to 0.746. Using majority voting as the gold standard consensus, this indicates that a single expert achieves approximately 65-75\% agreement with consensus judgments. \ourname's 80\% F1 score actually matches or exceeds typical individual human expert performance against consensus, demonstrating that the system has achieved human expert-level accuracy on this inherently subjective task. This subjectivity is fundamental to legal issue relevance assessment. Experts with identical training and rubrics often reach different legitimate conclusions on the same cases, reflecting interpretive differences rather than errors. This suggests that the realistic performance ceiling for this task is bounded by legitimate expert disagreement, likely around 75-85\% F1, and \ourname's performance places it within this realistic ceiling range. Thus, we argue that the gap to production deployment is therefore not primarily about improving accuracy beyond human expert levels, but rather about meeting the operational, robustness, and trust requirements of real-world legal practice, such as adding post-checking with real legal professionals after prediction.

\subsection{Diversity Discussion}
\label{app:diversity}
Table~\ref{tab:issue_gen} shows that the \emph{incremental} generator beats the one–pass \emph{baseline} on both quality and diversity.

First, for quality, Fréchet BERT Distance drops from 1311 to 1177 and
BERT Embedding Distance from 1354 to 1227. Because smaller distances mean the candidates sit nearer to the ground-truth issues in embedding space, these reductions (roughly 10\%) confirm that the staged prompts recover more of the key semantics that human experts mark.

On the diversity side, three signals move in the right direction. Self-EMBD rises from 211.8 to 225.1, so the candidates spread out further in representation space. Raw Self-BLEU nearly doubles at the 3–, 4–, and 5–gram levels (11.85/14.52/16.64 versus 24.90/30.36/34.95). If one prefers to report \(1-\)Self-BLEU, the gain is the same magnitude, just flipped in sign. Distinct-N climbs by about 40\%, adding more than three thousand new 3-, 4-, 5-grams to the pool.

In short, revealing facts to the LLM one step at a time both tightens semantic coverage and uncovers a broader set of legally plausible formulations—a combination that downstream filters and reviewers can exploit.

\subsection{Data Example: Case ID - IFSG681}
\label{app:case_example}
\subsubsection{Case Facts (Scenario)}
\begin{itemize}
    \item The developer, Sri Damansara Sdn Bhd, collected a booking fee of RM10,000 on 6 January 2012 prior to signing the sale and purchase agreement (SPA) dated 28 June 2012, which contravened Housing Development regulations.
    \item When the vacant possession was delivered on 22 December 2016, the purchasers filed a claim for damages for late delivery, calculating it from the booking fee payment date to the delivery date, which the Tribunal upheld.
    \item The developer argued that the calculation should start from the SPA date and questioned the validity of the Tribunal's decision and the method of calculating the purchase price considering a credit note provided.
\end{itemize}

\subsubsection{Ground Truth Issues}
\begin{enumerate}[label=\arabic*.]
    \item Whether the late delivery claim ought to be calculated from the expiry of 42 months from the date of the SPA of 28 June 2012 until the date of delivery of vacant possession.
    \item Whether the purchase price for the calculation of the late delivery claim should be based on the price as stated in the SPA or the reduced price via the device of a credit note issued by the developer.
\end{enumerate}

\subsubsection{Issue Candidates}

\paragraph{Scenario\#1:}
\begin{enumerate}[label=\arabic*.]
    \item The developer, Sri Damansara Sdn Bhd, collected a booking fee of RM10,000 on 6 January 2012 prior to signing the sale and purchase agreement (SPA) dated 28 June 2012, which contravened Housing Development regulations.
\end{enumerate}

\paragraph{Generated Issue Candidates Given Scenario\#1:}
\begin{enumerate}[label=\arabic*.]
    \item Whether the collection of a booking fee by Sri Damansara Sdn Bhd prior to signing the sale and purchase agreement (SPA) is in violation of Housing Development regulations.
    \item Whether the developer's action of collecting a booking fee on 6 January 2012, nearly six months before the SPA was signed on 28 June 2012, is legal and permissible under applicable housing laws.
\end{enumerate}

\paragraph{Scenario\#2:}
\begin{enumerate}[label=\arabic*.]
    \item The developer, Sri Damansara Sdn Bhd, collected a booking fee of RM10,000 on 6 January 2012 prior to signing the sale and purchase agreement (SPA) dated 28 June 2012, which contravened Housing Development regulations.
    \item When the vacant possession was delivered on 22 December 2016, the purchasers filed a claim for damages for late delivery, calculating it from the booking fee payment date to the delivery date, which the Tribunal upheld.
\end{enumerate}

\paragraph{Generated Issue Candidates Given Scenario\#2:}
\begin{enumerate}[label=\arabic*.]
    \item Whether the developer's collection of a booking fee prior to signing the sale and purchase agreement (SPA) was in contravention of Housing Development regulations.
    \item Whether the calculation of late delivery damages should commence from the date of the booking fee payment or from the date of the signed sale and purchase agreement (SPA).
\end{enumerate}

\paragraph{Scenario\#3:}
\begin{enumerate}[label=\arabic*.]
    \item The developer, Sri Damansara Sdn Bhd, collected a booking fee of RM10,000 on 6 January 2012 prior to signing the sale and purchase agreement (SPA) dated 28 June 2012, which contravened Housing Development regulations.
    \item When the vacant possession was delivered on 22 December 2016, the purchasers filed a claim for damages for late delivery, calculating it from the booking fee payment date to the delivery date, which the Tribunal upheld.
    \item The developer argued that the calculation should start from the SPA date and questioned the validity of the Tribunal's decision and the method of calculating the purchase price considering a credit note provided.
\end{enumerate}

\paragraph{Generated Issue Candidates Given Scenario\#3:}
\begin{enumerate}[label=\arabic*.]
    \item Whether the developer's collection of a booking fee prior to signing the Sale and Purchase Agreement (SPA) was in contravention of Housing Development regulations.
    \item Whether the calculation of damages for late delivery should start from the date of the booking fee payment or the date of the SPA.
    \item Whether the Tribunal's decision to uphold the purchasers' claim for damages based on the booking fee payment date is valid.
    \item Whether the method of calculating the purchase price should consider the credit note provided by the developer.
\end{enumerate}

\subsection{Example of Sparsity of Mutual Information}
\label{app:example_sparsity}
Legal Facts:
\begin{enumerate}
    \item Fact\#1. The Appellants, Tioh Chee Seng and Hew Fui Li, purchased a residential unit in Ayuman Suites under a sale and purchase agreement dated 21 October 2015.
    \item Fact\#2. The 1st Respondent, Talent Team Sdn. Bhd., is the developer responsible for constructing Ayuman Suites, while the 2nd Respondent, a legal firm, is alleged to be the stakeholder of certain sums related to the purchase.
    \item Fact\#3. The Appellants claimed late delivery of the property, seeking liquidated ascertained damages (LAD) based on delays exceeding the stipulated timeframes for completion.
    \item Fact\#4. The main dispute centers on the calculation of the delay—specifically, whether it should be based on the booking fee payment date or the date of signing the sales and purchase agreement.
    \item Fact\#5. The Sessions Court ruled that the LAD calculation should commence from the signing of the sales and purchase agreement.
\end{enumerate}

Identified Legal Issues (Selected):
\begin{enumerate}
    \item Issue \#1 (Identified when providing Facts 1–3): Whether the 2nd Respondent, as the alleged stakeholder, has any liability regarding the sums related to the purchase.
    \item Issue \#2 (Identified when providing Facts 1–4): Whether the legal firm, acting as the alleged stakeholder, bears any responsibility in the dispute over the late delivery and calculation of LAD.
    \item Issue \#3 (Identified when providing Facts 1–5): Whether the 2nd Respondent (legal firm) has any liability as the alleged stakeholder of certain sums related to the purchase.
\end{enumerate}

In this case, Issues \#1, \#2, and \#3 all relate to the liability of the 2nd Respondent (Legal Firm). Once Issue \#1 is identified based on Facts 1–3, the subsequent iterations—despite introducing additional facts—continue to surface the same or similar issues. This suggests that Facts \#4 and \#5 contribute little to the identification of this legal issue.

\subsection{Prompt Template for Incremental Issue Generation}
\label{app:incremental_issue_gen_prompt}
\begin{lstlisting}[style=prompt, caption={Prompt for Incremental Issue Generation}]
    Scenario: \{scenario\}

This scenario describes a legal case. Based on the details provided, please identify the most relevant legal issues.

Guidelines:
1.	Do not alter or deviate from the meaning presented in the scenario.
2.	Format each legal issue as "Whether ...", for example: "Whether the alleged agreement between the plaintiff and defendant is enforceable considering the Statute of Frauds."
3.	Provide your response strictly in JSON format as shown below:
    
{["YOUR FIRST LEGAL ISSUE","YOUR SECOND LEGAL ISSUE", ...}
\end{lstlisting}

\subsection{Evaluation Guideline for Human} 
\label{evaluationGuide}

\paragraph{Facts Evaluation}

\textbf{High Distinction (HD):}
\begin{itemize}
    \item Facts are presented clearly and concisely in a structured point form.
    \item Closely aligned with statutory language and terminology.
    \item No irrelevant details, and all essential elements are thoroughly included.
\end{itemize}

\textbf{Pass:}
\begin{itemize}
    \item Facts are mostly accurate and clear, though some minor details may be missing or imprecise.
    \item Minor elements could be better structured or clarified.
\end{itemize}

\textbf{Not Pass:}
\begin{itemize}
    \item Facts are incomplete, unclear, or contain irrelevant information that detracts from the analysis.
    \item Key details are missing, leading to a lack of proper context.
\end{itemize}

\textbf{Neutral:}
\begin{itemize}
    \item Facts are presented and generally acceptable, but lack the depth or clarity needed for proper evaluation.
    \item Facts may not align clearly with the case or legal standards, preventing detailed assessment.
\end{itemize}

\paragraph{Issues Evaluation}

\textbf{High Distinction (HD):}
\begin{itemize}
    \item All relevant legal issues are clearly identified in a structured manner, typically starting with \textit{"Whether..."}.
    \item Issues are aligned with the facts and the applicable rules, demonstrating a comprehensive understanding.
\end{itemize}

\textbf{Pass:}
\begin{itemize}
    \item Most key legal issues are identified, but some may be phrased imprecisely or omitted.
    \item Overall, the issues are reasonable, but there may be minor gaps in alignment with facts and rules.
\end{itemize}

\textbf{Not Pass:}
\begin{itemize}
    \item Significant legal issues are missing or misidentified, demonstrating a poor understanding of the case.
    \item Issues are formulated incorrectly or too broadly.
\end{itemize}

\textbf{Neutral:}
\begin{itemize}
    \item Issues are present, but lack clarity, structure, or alignment with the case, making it difficult to assess their relevance.
\end{itemize}

\subsection{Baseline Settings}
\label{app:hyperparameters}
We run all the experiments on NVIDIA A100 GPUs. For the hyperparameters of the LLMs, e.g., Temperature, Top-$p$, etc, we use the default settings for all commercial models, including Claude and \gpto. The random seeds are set to 42.

We adopt the default settings for all generative models. Note that for Qwen3-14B, we disable the thinking function and only look into their \texttt{no think} mode. For the regression methods, we use GridCV to tune the hyperparameters, including $\lambda$, C, etc. For the deep learning methods, we tune the learning rate and epoch based on the validation set.

\subsection{Prompt Template for Preliminary Results}
\label{app:prompt_preliminary}

\begin{lstlisting}[style=prompt, caption={Prompt for Preliminary Results}]
You are a meticulous judge deciding whether the issue below is "Relevant" or
"Irrelevant" under the exact definitions provided.

Definitions (use exactly):
- "Relevant":  The issue is not only related to the scenario, but is directly
  tied to the main dispute or core facts of the case. It goes beyond merely
  stating a basic legal principle or background fact.
- "Irrelevant": The issue is either unrelated to the scenario or is only a
  fundamental/basic statement that does not bear on the case's primary
  controversy.

Think silently and step-by-step.  Follow these internal reasoning steps:
  1. Identify the main dispute and the core facts from the Scenario Facts.
  2. Compare the candidate Issue to the dispute: does it address that dispute
     directly, or is it merely background/unrelated?
  3. Decide if the Issue adds substantive analysis beyond a basic legal truism.
  4. Conclude "Relevant" or "Irrelevant" based on the definitions.
- Do NOT reveal your reasoning or the steps above.
- After finishing your internal analysis, output **exactly one word**-
  either "Relevant" or "Irrelevant"-and nothing else.

### Scenario Facts
{facts}

### Issue
{issue}

### Instruction
First reason internally following the steps.  
Then output one word: Relevant OR Irrelevant

Your response:
\end{lstlisting}

\subsection{Full List of RQ2}
\label{app:full_rq2}
\begin{table*}[h]
  \centering
  \caption{Results of Deep Learning Methods}  
  \label{tab:deep_arch_results}
    \begin{tabular}{lcccc}
      \hline
      Methods & F1 & Acc. & Prec. & Rec. \\ 
      \hline
      FFN$_{\text{oss}}$  
        & $\text{56.82}_{\pm\text{3.96}}$  
        & $\text{57.90}_{\pm\text{4.79}}$  
        & $\text{60.90}_{\pm\text{3.02}}$  
        & $\text{61.96}_{\pm\text{3.18}}$ \\ 
      CNN$_{\text{oss}}$  
        & $\text{39.35}_{\pm\text{4.36}}$  
        & $\text{59.92}_{\pm\text{13.08}}$  
        & $\text{44.21}_{\pm\text{9.52}}$  
        & $\text{49.40}_{\pm\text{2.40}}$ \\ 
      Tranf.$_{\text{oss}}$  
        & $\text{61.72}_{\pm\text{2.83}}$  
        & $\text{63.05}_{\pm\text{3.83}}$  
        & $\text{65.22}_{\pm\text{1.64}}$  
        & $\text{66.85}_{\pm\text{2.22}}$ \\ 
      LSTM$_{\text{oss}}$  
        & $\text{47.89}_{\pm\text{4.70}}$  
        & $\text{64.15}_{\pm\text{4.70}}$  
        & $\text{51.08}_{\pm\text{10.29}}$  
        & $\text{51.45}_{\pm\text{1.85}}$ \\ 
      ResNet$_{\text{oss}}$  
        & $\text{60.47}_{\pm\text{4.59}}$  
        & $\text{65.57}_{\pm\text{4.90}}$  
        & $\text{60.80}_{\pm\text{5.09}}$  
        & $\text{60.74}_{\pm\text{4.31}}$ \\ 
      \hline
      FFN$_{\text{Phi}}$  
        & $\text{60.79}_{\pm\text{5.32}}$  
        & $\text{66.73}_{\pm\text{9.21}}$  
        & $\text{65.60}_{\pm\text{6.67}}$  
        & $\text{61.96}_{\pm\text{2.92}}$ \\ 
      CNN$_{\text{Phi}}$  
        & $\text{40.47}_{\pm\text{9.20}}$  
        & $\text{56.17}_{\pm\text{16.12}}$  
        & $\text{47.73}_{\pm\text{23.15}}$  
        & $\text{50.11}_{\pm\text{2.64}}$ \\ 
      Tranf.$_{\text{Phi}}$  
        & $\text{70.53}_{\pm\text{3.74}}$  
        & $\text{75.33}_{\pm\text{5.80}}$  
        & $\text{75.06}_{\pm\text{4.81}}$  
        & $\text{71.13}_{\pm\text{2.46}}$ \\ 
      LSTM$_{\text{Phi}}$  
        & $\text{45.36}_{\pm\text{8.84}}$  
        & $\text{59.24}_{\pm\text{13.10}}$  
        & $\text{52.97}_{\pm\text{3.18}}$  
        & $\text{51.78}_{\pm\text{2.14}}$ \\ 
      \text{\text{ResNet$_{\text{Phi}}$}}  
        & $\mathbf{\text{67.33}_{\pm\text{3.61}}}$  
        & $\text{72.48}_{\pm\text{4.19}}$  
        & $\text{68.84}_{\pm\text{4.89}}$  
        & $\text{67.33}_{\pm\text{3.40}}$ \\ 
      \hline
      FFN$_{\text{Qwen}}$  
        & $\text{75.65}_{\pm\text{2.57}}$  
        & $\text{79.29}_{\pm\text{2.43}}$  
        & $\text{76.10}_{\pm\text{2.84}}$  
        & $\text{75.57}_{\pm\text{2.47}}$ \\ 
      CNN$_{\text{Qwen}}$  
        & $\text{43.33}_{\pm\text{3.08}}$  
        & $\text{69.78}_{\pm\text{0.96}}$  
        & $\text{54.78}_{\pm\text{24.83}}$  
        & $\text{51.22}_{\pm\text{1.51}}$ \\ 
      \text{\text{Tranf.$_{\text{Qwen}}$}}  
        & $\mathbf{\text{75.44}_{\pm\text{1.82}}}$  
        & $\text{80.14}_{\pm\text{2.02}}$  
        & $\text{78.22}_{\pm\text{3.64}}$  
        & $\text{74.39}_{\pm\text{2.10}}$ \\ 
      LSTM$_{\text{Qwen}}$  
        & $\text{63.48}_{\pm\text{2.94}}$  
        & $\text{65.90}_{\pm\text{3.79}}$  
        & $\text{64.04}_{\pm\text{1.93}}$  
        & $\text{65.81}_{\pm\text{1.84}}$ \\ 
      ResNet$_{\text{Qwen}}$  
        & $\text{67.21}_{\pm\text{3.76}}$  
        & $\text{72.30}_{\pm\text{2.93}}$  
        & $\text{67.42}_{\pm\text{3.66}}$  
        & $\text{67.06}_{\pm\text{3.81}}$ \\ 
      \hline
    \end{tabular}%
\end{table*}

\begin{table*}[h]
  \centering
  \caption{Results of Feature Selection Methods}  
  \label{tab:llm_select_aggregates}
    \begin{tabular}{lcccc}
      \hline
      Methods & F1 & Acc. & Prec. & Rec. \\ 
      \hline
      L1Reg$_\text{Phi}$  
        & $\text{80.01}_{\pm\text{3.61}}$ 
        & $\text{83.34}_{\pm\text{3.06}}$ 
        & $\text{81.13}_{\pm\text{3.93}}$ 
        & $\text{79.32}_{\pm\text{3.45}}$ \\ 
      L1Reg$_{\text{oss}}$  
        & $\text{65.69}_{\pm\text{3.74}}$ 
        & $\text{72.05}_{\pm\text{3.29}}$ 
        & $\text{66.91}_{\pm\text{4.30}}$ 
        & $\text{65.07}_{\pm\text{3.44}}$ \\ 
      L1Reg$_{\mathrm{Qwen}}$  
        & $\text{73.80}_{\pm\text{2.25}}$ 
        & $\text{78.03}_{\pm\text{2.04}}$ 
        & $\text{74.47}_{\pm\text{2.57}}$ 
        & $\text{73.39}_{\pm\text{2.22}}$ \\ 
      \hline
      L1SVC$_\text{Phi}$  
        & $\text{77.60}_{\pm\text{2.58}}$ 
        & $\text{80.89}_{\pm\text{1.97}}$ 
        & $\text{77.68}_{\pm\text{2.23}}$ 
        & $\text{77.62}_{\pm\text{2.93}}$ \\ 
      L1SVC$_{\mathrm{Qwen}}$  
        & $\text{70.61}_{\pm\text{2.59}}$ 
        & $\text{74.83}_{\pm\text{2.51}}$ 
        & $\text{70.68}_{\pm\text{2.77}}$ 
        & $\text{70.61}_{\pm\text{2.39}}$ \\ 
      L1SVC$_{\text{oss}}$  
        & $\text{64.38}_{\pm\text{1.57}}$ 
        & $\text{68.85}_{\pm\text{1.47}}$ 
        & $\text{64.20}_{\pm\text{1.57}}$ 
        & $\text{64.78}_{\pm\text{1.69}}$ \\ 
      LMSel.        & $39.10_{\pm0.49}$ & $64.23_{\pm1.34}$ & $33.73_{\pm0.22}$ & $46.52_{\pm0.99}$ \\
      LMSel.$_\text{L}$ & $45.63_{\pm1.79}$ & $50.25_{\pm1.72}$ & $46.27_{\pm1.76}$ & $45.84_{\pm1.92}$ \\ 
              \hline
      \multicolumn{5}{c}{LMSel.$^\text{P}$ + Regression (Continuous)}\\
      \hline
      
      Logistic                       & $74.86_{\pm2.85}$ & $77.44_{\pm2.71}$ & $74.24_{\pm2.71}$ & $76.32_{\pm2.85}$ \\
      SVC                            & $73.98_{\pm3.48}$ & $76.85_{\pm3.06}$ & $73.41_{\pm3.31}$ & $75.14_{\pm3.66}$ \\
      LDA                            & $71.45_{\pm3.35}$ & $77.19_{\pm2.59}$ & $73.75_{\pm3.52}$ & $70.36_{\pm3.19}$ \\
      Ridge                          & $73.95_{\pm2.31}$ & $76.60_{\pm2.09}$ & $73.31_{\pm2.17}$ & $75.41_{\pm2.46}$ \\
      \hline
      \multicolumn{5}{c}{LMSel.$^\text{P}$ + Regression (Binary)}\\
      \hline
      \uline{Logistic}      & $\text{57.09}_{\pm\text{1.55}}$ & $\text{64.90}_{\pm\text{1.43}}$ & $\text{57.79}_{\pm\text{1.41}}$ & $\text{57.06}_{\pm\text{1.48}}$ \\
\uline{SVC}           & $\text{57.78}_{\pm\text{1.57}}$ & $\text{66.33}_{\pm\text{0.82}}$ & $\text{58.90}_{\pm\text{1.20}}$ & $\text{57.64}_{\pm\text{1.52}}$ \\
\uline{LDA}           & $\text{53.23}_{\pm\text{2.31}}$ & $\text{65.74}_{\pm\text{2.19}}$ & $\text{56.38}_{\pm\text{3.18}}$ & $\text{54.07}_{\pm\text{1.77}}$ \\
\uline{Ridge}         & $\text{56.15}_{\pm\text{0.43}}$ & $\text{64.39}_{\pm\text{1.88}}$ & $\text{57.08}_{\pm\text{0.91}}$ & $\text{56.17}_{\pm\text{0.38}}$ \\
\hline
    \end{tabular}%
  
\end{table*}
\begin{table*}[h]
\footnotesize
  \centering
  \caption{Results of Regression Methods}  
  \label{tab:aggregated_results}
    \begin{tabular}{lcccc}
      \hline
      Methods & F1 & Acc. & Prec. & Rec. \\ 
      \hline
      \text{LR$_\text{Phi}$}           
        & $\text{79.70}_{\pm\text{2.93}}$ 
        & $\text{82.49}_{\pm\text{2.41}}$ 
        & $\text{79.58}_{\pm\text{2.89}}$ 
        & $\text{80.05}_{\pm\text{3.23}}$ \\ 
      \text{\uline{LR}$_\text{Phi}$}           
        & $\text{64.70}_{\pm\text{3.18}}$ 
        & $\text{68.86}_{\pm\text{2.81}}$ 
        & $\text{64.56}_{\pm\text{3.01}}$ 
        & $\text{65.41}_{\pm\text{3.51}}$ \\ 
      \text{\text{SVC$_\text{Phi}$}}  
        & $\text{\text{80.19}}_{\pm\text{2.83}}$ 
        & $\text{82.66}_{\pm\text{2.38}}$ 
        & $\text{79.67}_{\pm\text{2.70}}$ 
        & $\text{\text{81.01}}_{\pm\text{3.13}}$ \\ 
      \text{\uline{SVC}$_\text{Phi}$}  
        & $\text{62.30}_{\pm\text{3.56}}$ 
        & $\text{67.59}_{\pm\text{2.30}}$ 
        & $\text{62.39}_{\pm\text{3.18}}$ 
        & $\text{62.70}_{\pm\text{4.07}}$ \\ 
      \text{Ridge$_\text{Phi}$}        
        & $\text{80.10}_{\pm\text{2.86}}$ 
        & $\text{82.91}_{\pm\text{2.41}}$ 
        & $\text{80.06}_{\pm\text{2.89}}$ 
        & $\text{80.28}_{\pm\text{3.05}}$ \\ 
      \text{\uline{Ridge}$_\text{Phi}$}        
        & $\text{61.63}_{\pm\text{2.79}}$ 
        & $\text{65.99}_{\pm\text{1.87}}$ 
        & $\text{61.56}_{\pm\text{2.65}}$ 
        & $\text{62.44}_{\pm\text{3.47}}$ \\ 
      \text{\text{LDA$_\text{Phi}$}}  
        & $\text{79.56}_{\pm\text{4.01}}$ 
        & $\text{\text{83.50}}_{\pm\text{2.69}}$ 
        & $\text{\text{81.77}}_{\pm\text{2.97}}$ 
        & $\text{78.39}_{\pm\text{4.30}}$ \\ 
      \text{\text{\uline{LDA}$_\text{Phi}$}}  
        & $\text{63.83}_{\pm\text{3.09}}$ 
        & $\text{72.73}_{\pm\text{1.73}}$ 
        & $\text{67.73}_{\pm\text{2.41}}$ 
        & $\text{63.11}_{\pm\text{2.79}}$ \\ 
      SGD$_\text{Phi}$                   
        & $\text{73.90}_{\pm\text{9.33}}$ 
        & $\text{77.45}_{\pm\text{8.89}}$ 
        & $\text{74.68}_{\pm\text{9.97}}$ 
        & $\text{73.85}_{\pm\text{9.07}}$ \\ 
      \uline{SGD}$_\text{Phi}$                   
        & $\text{62.95}_{\pm\text{1.99}}$ 
        & $\text{67.09}_{\pm\text{2.54}}$ 
        & $\text{62.80}_{\pm\text{2.00}}$ 
        & $\text{63.58}_{\pm\text{1.57}}$ \\
      \text{RF$_\text{Phi}$}           
        & $\text{66.30}_{\pm\text{3.01}}$ 
        & $\text{75.76}_{\pm\text{2.49}}$ 
        & $\text{74.10}_{\pm\text{4.82}}$ 
        & $\text{64.99}_{\pm\text{2.54}}$ \\ 
      \text{\uline{RF}$_\text{Phi}$}           
        & $\text{61.56}_{\pm\text{4.09}}$ 
        & $\text{72.14}_{\pm\text{2.16}}$ 
        & $\text{66.97}_{\pm\text{4.58}}$ 
        & $\text{61.16}_{\pm\text{3.28}}$ \\ 
      KNN$_\text{Phi}$                   
        & $\text{66.94}_{\pm\text{3.24}}$ 
        & $\text{73.57}_{\pm\text{3.05}}$ 
        & $\text{69.04}_{\pm\text{4.09}}$ 
        & $\text{66.10}_{\pm\text{3.02}}$ \\ 
      \uline{KNN}$_\text{Phi}$                   
        & $\text{62.46}_{\pm\text{3.15}}$ 
        & $\text{71.63}_{\pm\text{2.75}}$ 
        & $\text{66.25}_{\pm\text{4.37}}$ 
        & $\text{61.78}_{\pm\text{2.77}}$ \\ 
      DT$_\text{Phi}$                    
        & $\text{62.52}_{\pm\text{2.13}}$ 
        & $\text{68.18}_{\pm\text{1.54}}$ 
        & $\text{62.65}_{\pm\text{1.99}}$ 
        & $\text{62.51}_{\pm\text{2.36}}$ \\ 
      \uline{DT}$_\text{Phi}$                    
        & $\text{60.42}_{\pm\text{4.01}}$ 
        & $\text{65.24}_{\pm\text{3.80}}$ 
        & $\text{60.50}_{\pm\text{3.76}}$ 
        & $\text{60.98}_{\pm\text{4.18}}$ \\ 
      NB$_\text{Phi}$                    
        & $\text{49.61}_{\pm\text{3.17}}$ 
        & $\text{51.00}_{\pm\text{4.18}}$ 
        & $\text{53.17}_{\pm\text{2.32}}$ 
        & $\text{53.52}_{\pm\text{2.62}}$ \\ 
      \uline{NB}$_\text{Phi}$                    
        & $\text{39.63}_{\pm\text{4.76}}$ 
        & $\text{61.12}_{\pm\text{13.55}}$ 
        & $\text{46.36}_{\pm\text{6.86}}$ 
        & $\text{49.69}_{\pm\text{0.55}}$ \\ 
      \hline
      LR$_\text{Qwen}$                   
        & $\text{77.13}_{\pm\text{1.35}}$ 
        & $\text{80.55}_{\pm\text{1.13}}$ 
        & $\text{77.37}_{\pm\text{1.42}}$ 
        & $\text{77.01}_{\pm\text{1.51}}$ \\ 
      \uline{LR}$_\text{Qwen}$                   
        & $\text{76.26}_{\pm\text{3.03}}$ 
        & $\text{79.71}_{\pm\text{3.07}}$ 
        & $\text{76.60}_{\pm\text{3.63}}$ 
        & $\text{76.17}_{\pm\text{2.47}}$ \\ 
      SVC$_\text{Qwen}$                  
        & $\text{74.80}_{\pm\text{1.47}}$ 
        & $\text{79.04}_{\pm\text{0.95}}$ 
        & $\text{75.73}_{\pm\text{1.17}}$ 
        & $\text{74.27}_{\pm\text{1.89}}$ \\ 
      \uline{SVC}$_\text{Qwen}$                  
        & $\text{74.14}_{\pm\text{3.77}}$ 
        & $\text{78.11}_{\pm\text{3.69}}$ 
        & $\text{74.76}_{\pm\text{4.46}}$ 
        & $\text{73.81}_{\pm\text{3.23}}$ \\ 
      Ridge$_\text{Qwen}$                
        & $\text{74.67}_{\pm\text{4.01}}$ 
        & $\text{78.36}_{\pm\text{3.95}}$ 
        & $\text{75.06}_{\pm\text{4.66}}$ 
        & $\text{74.52}_{\pm\text{3.39}}$ \\ 
      \uline{Ridge}$_\text{Qwen}$                
        & $\text{73.17}_{\pm\text{3.32}}$ 
        & $\text{77.18}_{\pm\text{3.34}}$ 
        & $\text{73.63}_{\pm\text{3.88}}$ 
        & $\text{72.99}_{\pm\text{2.84}}$ \\ 
      \text{LDA$_\text{Qwen}$}          
        & $\text{76.46}_{\pm\text{3.00}}$ 
        & $\text{79.96}_{\pm\text{2.85}}$ 
        & $\text{76.79}_{\pm\text{3.57}}$ 
        & $\text{76.27}_{\pm\text{2.59}}$ \\ 
      \text{\uline{LDA}$_\text{Qwen}$}          
        & $\text{76.75}_{\pm\text{3.01}}$ 
        & $\text{80.30}_{\pm\text{2.87}}$ 
        & $\text{77.26}_{\pm\text{3.69}}$ 
        & $\text{76.45}_{\pm\text{2.62}}$ \\ 
      \text{SGD$_\text{Qwen}$}          
        & $\text{75.24}_{\pm\text{2.94}}$ 
        & $\text{79.37}_{\pm\text{3.08}}$ 
        & $\text{76.65}_{\pm\text{4.39}}$ 
        & $\text{74.58}_{\pm\text{2.28}}$ \\ 
      \text{\uline{SGD}$_\text{Qwen}$}          
        & $\text{73.13}_{\pm\text{4.49}}$ 
        & $\text{78.61}_{\pm\text{2.94}}$ 
        & $\text{75.57}_{\pm\text{3.71}}$ 
        & $\text{72.06}_{\pm\text{4.44}}$ \\ 
      \text{RF$_\text{Qwen}$}          
        & $\text{74.45}_{\pm\text{2.94}}$ 
        & $\text{79.63}_{\pm\text{2.74}}$ 
        & $\text{77.52}_{\pm\text{4.56}}$ 
        & $\text{73.04}_{\pm\text{2.42}}$ \\
      \text{\uline{RF}$_\text{Qwen}$}          
        & $\text{73.55}_{\pm\text{2.53}}$ 
        & $\text{78.78}_{\pm\text{2.90}}$ 
        & $\text{76.97}_{\pm\text{5.40}}$ 
        & $\text{72.36}_{\pm\text{2.07}}$ \\
      \text{KNN$_\text{Qwen}$}          
        & $\text{74.53}_{\pm\text{2.06}}$ 
        & $\text{79.12}_{\pm\text{1.81}}$ 
        & $\text{76.06}_{\pm\text{2.61}}$ 
        & $\text{73.66}_{\pm\text{2.01}}$ \\ 
      \text{\uline{KNN}$_\text{Qwen}$}          
        & $\text{74.35}_{\pm\text{1.58}}$ 
        & $\text{79.12}_{\pm\text{1.45}}$ 
        & $\text{76.16}_{\pm\text{2.16}}$ 
        & $\text{73.36}_{\pm\text{1.52}}$ \\ 
      \text{DT$_\text{Qwen}$}           
        & $\text{67.81}_{\pm\text{4.88}}$ 
        & $\text{71.79}_{\pm\text{5.57}}$ 
        & $\text{68.72}_{\pm\text{4.76}}$ 
        & $\text{68.67}_{\pm\text{4.08}}$ \\ 
      \text{\uline{DT}$_\text{Qwen}$}           
        & $\text{67.83}_{\pm\text{3.37}}$ 
        & $\text{72.30}_{\pm\text{3.03}}$ 
        & $\text{68.14}_{\pm\text{3.54}}$ 
        & $\text{68.20}_{\pm\text{3.60}}$ \\ 
      \text{NB$_\text{Qwen}$}           
        & $\text{65.53}_{\pm\text{4.02}}$ 
        & $\text{69.61}_{\pm\text{3.86}}$ 
        & $\text{65.49}_{\pm\text{3.86}}$ 
        & $\text{66.24}_{\pm\text{3.93}}$ \\ 
      \text{\uline{NB}$_\text{Qwen}$}           
        & $\text{65.50}_{\pm\text{3.83}}$ 
        & $\text{69.44}_{\pm\text{3.99}}$ 
        & $\text{65.49}_{\pm\text{3.67}}$ 
        & $\text{66.27}_{\pm\text{3.56}}$ \\
      \hline
      LR$_\text{oss}$                    
        & $\text{68.33}_{\pm\text{1.54}}$ 
        & $\text{72.56}_{\pm\text{1.50}}$ 
        & $\text{68.16}_{\pm\text{1.66}}$ 
        & $\text{68.59}_{\pm\text{1.41}}$ \\ 
      \uline{LR}$_\text{oss}$                    
        & $\text{65.06}_{\pm\text{2.24}}$ 
        & $\text{69.87}_{\pm\text{1.51}}$ 
        & $\text{65.10}_{\pm\text{2.03}}$ 
        & $\text{65.39}_{\pm\text{2.71}}$ \\ 
      SVC$_\text{oss}$                   
        & $\text{66.38}_{\pm\text{2.89}}$ 
        & $\text{71.80}_{\pm\text{2.54}}$ 
        & $\text{66.79}_{\pm\text{3.06}}$ 
        & $\text{66.09}_{\pm\text{2.79}}$ \\ 
      \uline{SVC}$_\text{oss}$                   
        & $\text{61.36}_{\pm\text{2.14}}$ 
        & $\text{66.83}_{\pm\text{1.71}}$ 
        & $\text{61.45}_{\pm\text{2.11}}$ 
        & $\text{61.54}_{\pm\text{2.30}}$ \\ 
      Ridge$_\text{oss}$                 
        & $\text{67.08}_{\pm\text{2.33}}$ 
        & $\text{71.38}_{\pm\text{1.94}}$ 
        & $\text{67.00}_{\pm\text{2.20}}$ 
        & $\text{67.53}_{\pm\text{2.69}}$ \\ 
      \uline{Ridge}$_\text{oss}$                 
        & $\text{62.23}_{\pm\text{1.53}}$ 
        & $\text{66.58}_{\pm\text{0.84}}$ 
        & $\text{62.11}_{\pm\text{1.38}}$ 
        & $\text{62.93}_{\pm\text{1.98}}$ \\ 
      \text{\text{LDA$_\text{oss}$}}   
        & $\text{68.15}_{\pm\text{1.77}}$ 
        & $\mathbf{\text{73.32}_{\pm\text{2.81}}}$ 
        & $\text{\text{69.39}}_{\pm\text{3.20}}$ 
        & $\text{\text{67.95}}_{\pm\text{1.25}}$ \\ 
      \text{\text{\uline{LDA}$_\text{oss}$}}   
        & $\text{65.83}_{\pm\text{2.17}}$ 
        & $\text{70.71}_{\pm\text{2.08}}$ 
        & $\text{65.90}_{\pm\text{2.28}}$ 
        & $\text{65.91}_{\pm\text{2.14}}$ \\ 
      SGD$_\text{oss}$                   
        & $\text{65.10}_{\pm\text{3.68}}$ 
        & $\text{71.30}_{\pm\text{4.12}}$ 
        & $\text{66.49}_{\pm\text{5.20}}$ 
        & $\text{64.53}_{\pm\text{3.12}}$ \\ 
      \uline{SGD}$_\text{oss}$                   
        & $\text{63.40}_{\pm\text{1.37}}$ 
        & $\text{67.59}_{\pm\text{2.22}}$ 
        & $\text{63.52}_{\pm\text{1.73}}$ 
        & $\text{64.17}_{\pm\text{1.49}}$ \\ 
      \text{RF$_\text{oss}$}            
        & $\text{63.28}_{\pm\text{3.80}}$ 
        & $\text{73.32}_{\pm\text{2.13}}$ 
        & $\text{69.17}_{\pm\text{3.86}}$ 
        & $\text{62.62}_{\pm\text{3.20}}$ \\ 
      \text{\uline{RF}$_\text{oss}$}            
        & $\text{62.26}_{\pm\text{2.59}}$ 
        & $\text{72.56}_{\pm\text{1.18}}$ 
        & $\text{67.69}_{\pm\text{1.97}}$ 
        & $\text{61.69}_{\pm\text{2.27}}$ \\ 
      KNN$_\text{oss}$                   
        & $\text{65.03}_{\pm\text{1.51}}$ 
        & $\text{72.81}_{\pm\text{1.26}}$ 
        & $\text{67.92}_{\pm\text{2.06}}$ 
        & $\text{64.21}_{\pm\text{1.51}}$ \\ 
      \uline{KNN}$_\text{oss}$                   
        & $\text{63.01}_{\pm\text{1.38}}$ 
        & $\text{70.79}_{\pm\text{0.94}}$ 
        & $\text{65.11}_{\pm\text{1.13}}$ 
        & $\text{62.52}_{\pm\text{1.52}}$ \\ 
      DT$_\text{oss}$                    
        & $\text{58.19}_{\pm\text{2.68}}$ 
        & $\text{61.88}_{\pm\text{3.74}}$ 
        & $\text{58.97}_{\pm\text{2.67}}$ 
        & $\text{59.71}_{\pm\text{2.83}}$ \\ 
      \uline{DT}$_\text{oss}$                    
        & $\text{61.29}_{\pm\text{2.84}}$ 
        & $\text{65.24}_{\pm\text{2.50}}$ 
        & $\text{61.06}_{\pm\text{2.74}}$ 
        & $\text{62.08}_{\pm\text{3.03}}$ \\ 
      NB$_\text{oss}$                    
        & $\text{42.13}_{\pm\text{3.73}}$ 
        & $\text{43.01}_{\pm\text{3.32}}$ 
        & $\text{57.39}_{\pm\text{3.48}}$ 
        & $\text{55.50}_{\pm\text{2.65}}$ \\ 
      \uline{NB}$_\text{oss}$                    
        & $\text{41.73}_{\pm\text{3.60}}$ 
        & $\text{42.34}_{\pm\text{3.13}}$ 
        & $\text{54.47}_{\pm\text{2.76}}$ 
        & $\text{53.74}_{\pm\text{2.22}}$ \\ 
      \hline
    \end{tabular}%
\end{table*}

\section{RQ3: Selection Stability Analysis}
\label{app:rq3_stability}

\paragraph{Methodology}
\label{app:l1_stability_method}

\subpara{Bootstrap Stability Selection Protocol}
For both L1 Logistic Regression and L1 SVC, we employ bootstrap-based stability selection within nested 5-fold cross-validation.

\textbf{Outer loop (5-fold CV):} Split data into 5 folds. For each fold $k$: (i) Use 4 folds for training, 1 fold for testing; (ii) Perform hyperparameter tuning via inner 5-fold CV on training data; (iii) Select optimal regularization parameter $C_k$; (iv) Run stability selection procedure; (v) Evaluate on held-out test fold.

\textbf{Stability selection within each training fold:} Perform 100 bootstrap iterations. In each iteration $i$: randomly subsample 60\% of training data (without replacement), fit L1 model using optimal $C_k$ on subsample, and record which features have non-zero coefficients: $S_i = \{j : w_j \neq 0\}$. Then compute selection frequency for each feature: $\text{freq}(j) = \frac{1}{100}\sum_{i=1}^{100} \mathbb{1}[j \in S_i]$. Apply threshold $\tau \in \{0.3, 0.4, 0.5, 0.6\}$: feature $j$ is "stable" in fold $k$ if $\text{freq}(j) \geq \tau$.

\textbf{Cross-fold consistency analysis:} For each feature $j$, count in how many folds it was deemed stable: $\text{stability}(j) = \sum_{k=1}^{5} \mathbb{1}[\text{freq}_k(j) \geq \tau]$. We categorize features as: always selected ($\text{stability}(j) = 5$), highly stable ($\text{stability}(j) = 4$), moderately stable ($\text{stability}(j) = 3$), unstable ($\text{stability}(j) \in \{1, 2\}$), and never selected ($\text{stability}(j) = 0$).

\subpara{Comparison with Standard L2 Linear Models}
To assess whether L1-selected features are genuinely important, we fit Standard (L2-regularized) Logistic Regression on the full dataset using optimal $C$ determined by cross-validation. This provides reference coefficients $\beta_j^{\text{L2}}$ and p-values $p_j$ for all 2464 features. We then compare L1-selected features' L1 coefficients versus Standard LR coefficients. Our hypothesis is that if L1 identifies truly important features, $|\beta_j^{\text{L1}}|$ should correlate positively with $|\beta_j^{\text{L2}}|$. However, we observe near-zero or negative correlations, proving disagreement.

\paragraph{L1 Logistic Regression Detailed Results}
\label{app:l1_lr_results}

\subpara{Complete Selection Statistics}
Table~\ref{tab:l1_lr_full_stats} presents comprehensive statistics across all four threshold configurations.

\begin{table*}[h]
\centering
\caption{Complete L1 LR Selection Statistics}
\label{tab:l1_lr_full_stats}
\small
\begin{tabular}{lrrrr}
\toprule
\textbf{Metric} & \textbf{0.3} & \textbf{0.4} & \textbf{0.5} & \textbf{0.6} \\
\midrule
\multicolumn{5}{l}{\textit{Selection Consistency Across Folds}} \\
Always (5/5 folds) & 13 (0.53\%) & 5 (0.20\%) & 2 (0.08\%) & 1 (0.04\%) \\
Highly stable (4/5) & 30 (1.22\%) & 15 (0.61\%) & 4 (0.16\%) & 2 (0.08\%) \\
Moderately stable (3/5) & 71 (2.88\%) & 24 (0.97\%) & 18 (0.73\%) & 6 (0.24\%) \\
Unstable (1-2/5) & 519 (21.1\%) & 326 (13.2\%) & 193 (7.8\%) & 106 (4.3\%) \\
Never selected & 1831 (74.3\%) & 2094 (85.0\%) & 2247 (91.2\%) & 2349 (95.3\%) \\
\midrule
\multicolumn{5}{l}{\textit{Performance Metrics}} \\
Mean F1 & 0.736±0.021 & 0.722±0.022 & 0.729±0.053 & 0.699±0.046 \\
Mean Accuracy & 0.775±0.010 & 0.758±0.020 & 0.727±0.033 & 0.726±0.046 \\
\midrule
\multicolumn{5}{l}{\textit{Per-Fold Feature Count Variation}} \\
Mean \# features & 211.8 & 113.2 & 61.6 & 31.0 \\
Std \# features & 129.2 & 73.3 & 43.3 & 21.2 \\
Min features & 35 & 21 & 11 & 8 \\
Max features & 362 & 192 & 112 & 65 \\
Range (max/min) & 10.3× & 9.1× & 10.2× & 8.1× \\
\midrule
\multicolumn{5}{l}{\textit{Hyperparameter Statistics}} \\
C range & 0.17-100 & 0.17-100 & 0.17-100 & 0.17-100 \\
C ratio (max/min) & 599× & 599× & 599× & 599× \\
\midrule
\multicolumn{5}{l}{\textit{Correlation with Standard LR}} \\
L1-L2 coefficient corr. & -0.026 & -0.059 & -0.106 & -0.249 \\
\bottomrule
\end{tabular}
\end{table*}

\subpara{Always-Selected Features Analysis}
\label{app:l1_lr_stable_features}
Table~\ref{tab:l1_lr_always_selected} examines all features selected in all 5 folds for each threshold configuration, comparing their L1 coefficients with Standard LR weights.

\begin{table*}[h]
\centering
\caption{L1 LR Always-Selected Features: Comparison with Standard LR}
\label{tab:l1_lr_always_selected}
\begin{tabular}{clrrrr}
\toprule
\textbf{Threshold} & \textbf{Feature} & \textbf{L1 $\beta$} & \textbf{L1 SD} & \textbf{Std LR $\beta$} & \textbf{p-value} \\
\midrule
0.6 & f1914 & 1.034 & 0.902 & 0.061 & 0.9997 \\
\midrule
\multirow{2}{*}{0.5}
& f1914 & 1.034 & 0.902 & 0.061 & 0.9997 \\
& f2076 & 0.409 & 0.537 & 0.051 & 0.9997 \\
\midrule
\multirow{5}{*}{0.4}
& f1914 & 1.034 & 0.902 & 0.061 & 0.9997 \\
& f1534 & -0.936 & 0.985 & -0.061 & 0.9998 \\
& f1972 & 0.703 & 0.455 & 0.065 & 0.9998 \\
& f1980 & -0.610 & 0.640 & -0.059 & 0.9998 \\
& f2076 & 0.409 & 0.537 & 0.051 & 0.9997 \\
\midrule
\multirow{13}{*}{0.3}
& f1914 & 1.034 & 0.902 & 0.061 & 0.9997 \\
& f1534 & -0.936 & 0.985 & -0.061 & 0.9998 \\
& f598 & 0.775 & 0.531 & 0.044 & 0.9998 \\
& f1989 & 0.704 & 0.654 & 0.041 & 0.9998 \\
& f1972 & 0.703 & 0.455 & 0.065 & 0.9998 \\
& f1821 & -0.692 & 0.744 & -0.053 & 0.9998 \\
& f1784 & 0.640 & 0.491 & 0.057 & 0.9998 \\
& f1980 & -0.610 & 0.640 & -0.059 & 0.9998 \\
& f440 & 0.608 & 0.518 & 0.061 & 0.9998 \\
& f2076 & 0.409 & 0.537 & 0.051 & 0.9997 \\
& f126 & 0.268 & 0.310 & 0.054 & 0.9998 \\
& f1960 & -0.255 & 0.436 & -0.048 & 0.9998 \\
& f2352 & -0.004 & 0.009 & 0.011 & 0.9999 \\
\bottomrule
\end{tabular}
\end{table*}

\begin{figure*}
    \centering
    \includegraphics[width=0.8\linewidth]{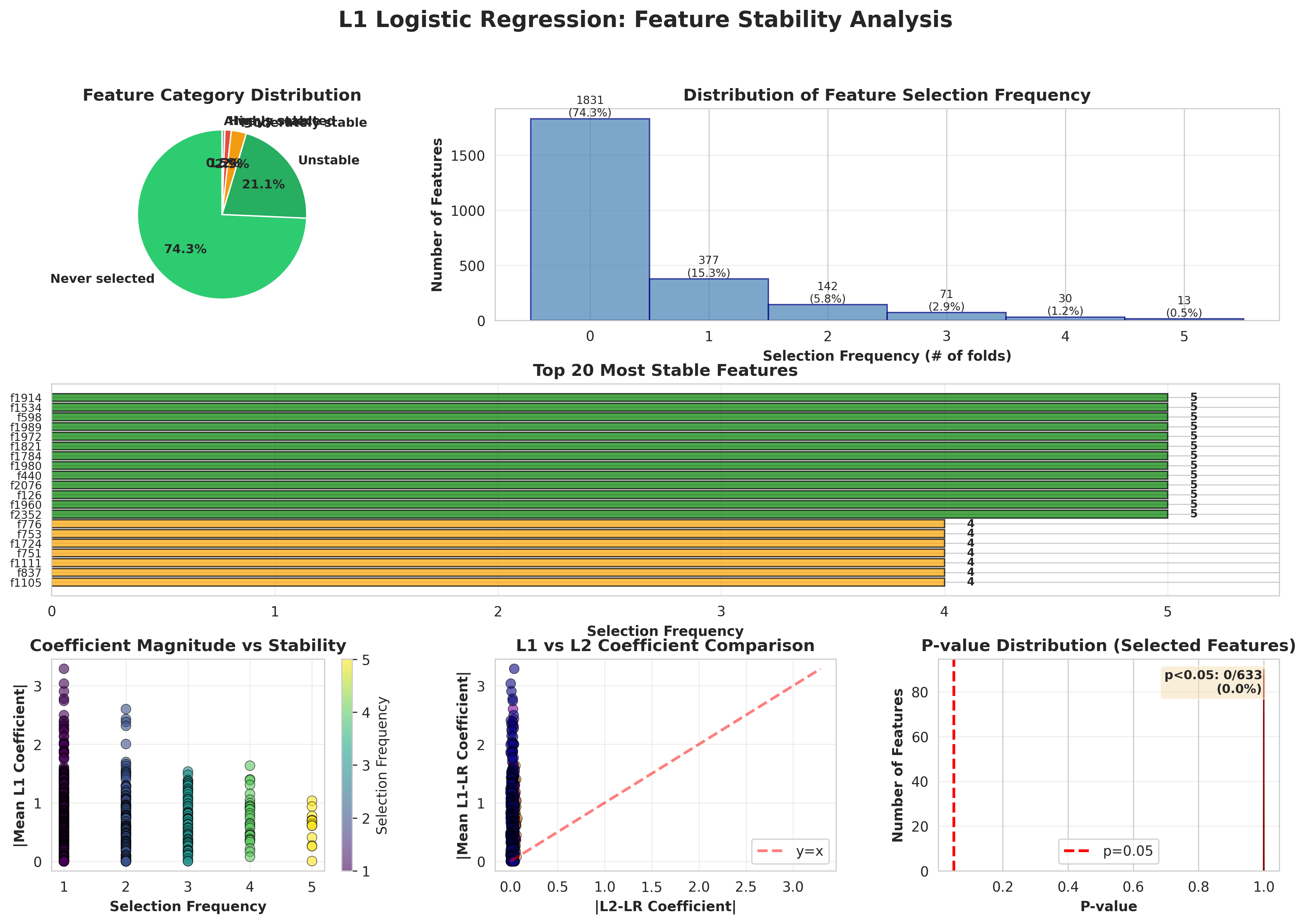}
    \caption{Stability Overview of L1 Regression (Threshold 0.3)}
    \label{fig:l1_lr_stability_03}
\end{figure*}

\begin{figure*}
    \centering
    \includegraphics[width=0.8\linewidth]{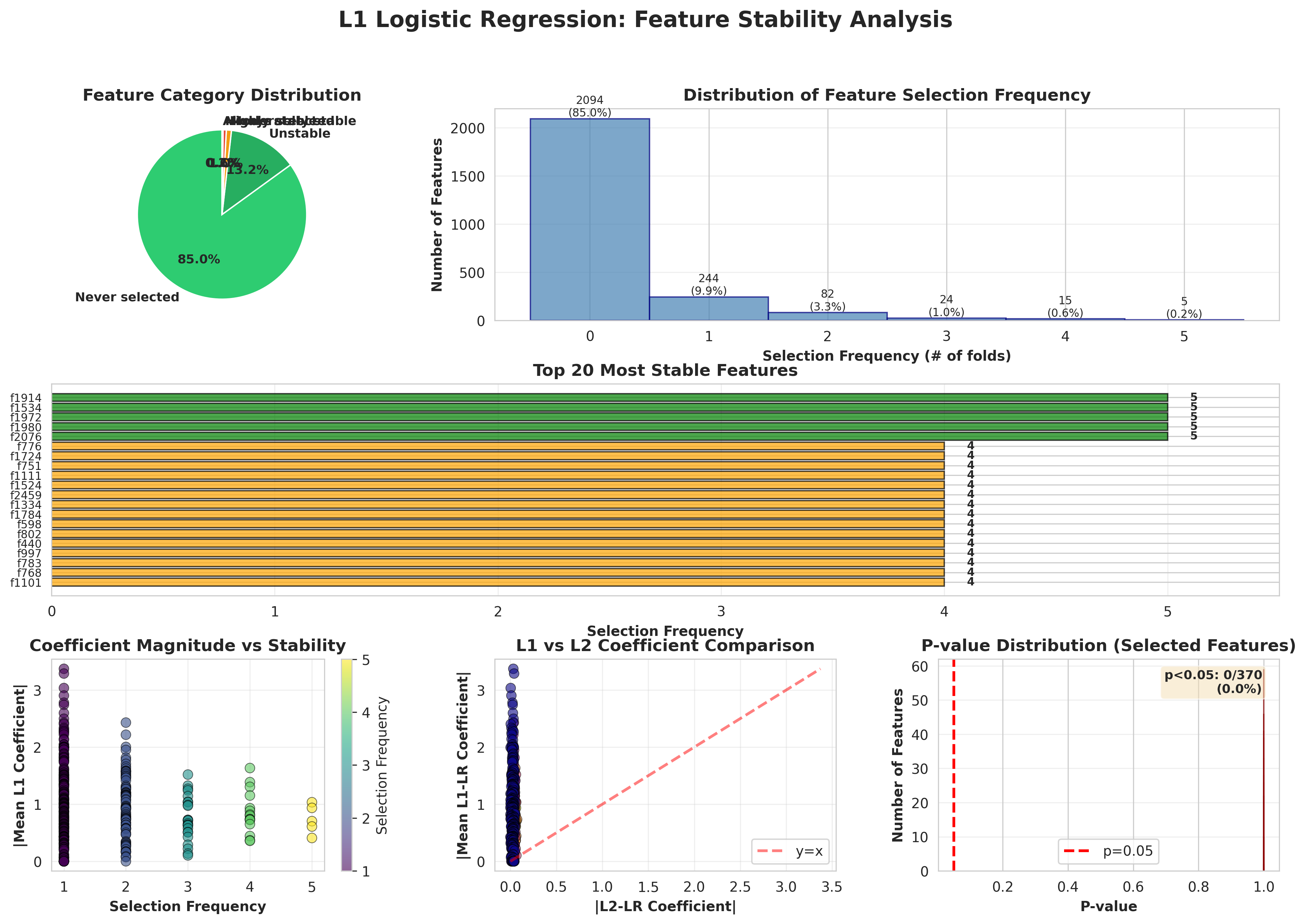}
    \caption{Stability Overview of L1 Regression (Threshold 0.4)}
    \label{fig:l1_lr_stability_04}
\end{figure*}

\begin{figure*}
    \centering
    \includegraphics[width=0.8\linewidth]{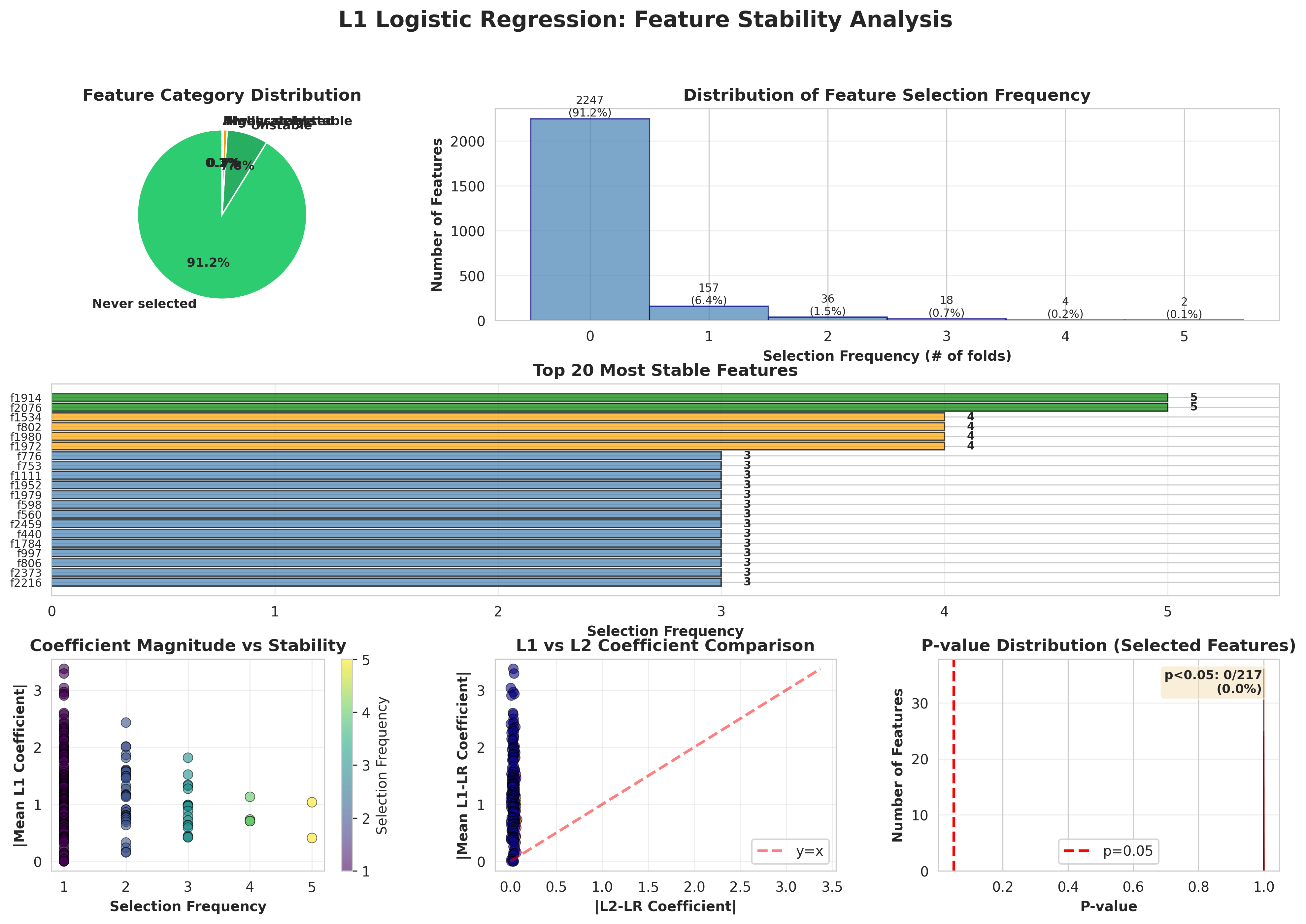}
    \caption{Stability Overview of L1 Regression (Threshold 0.5)}
    \label{fig:l1_lr_stability_05}
\end{figure*}

\begin{figure*}
    \centering
    \includegraphics[width=0.8\linewidth]{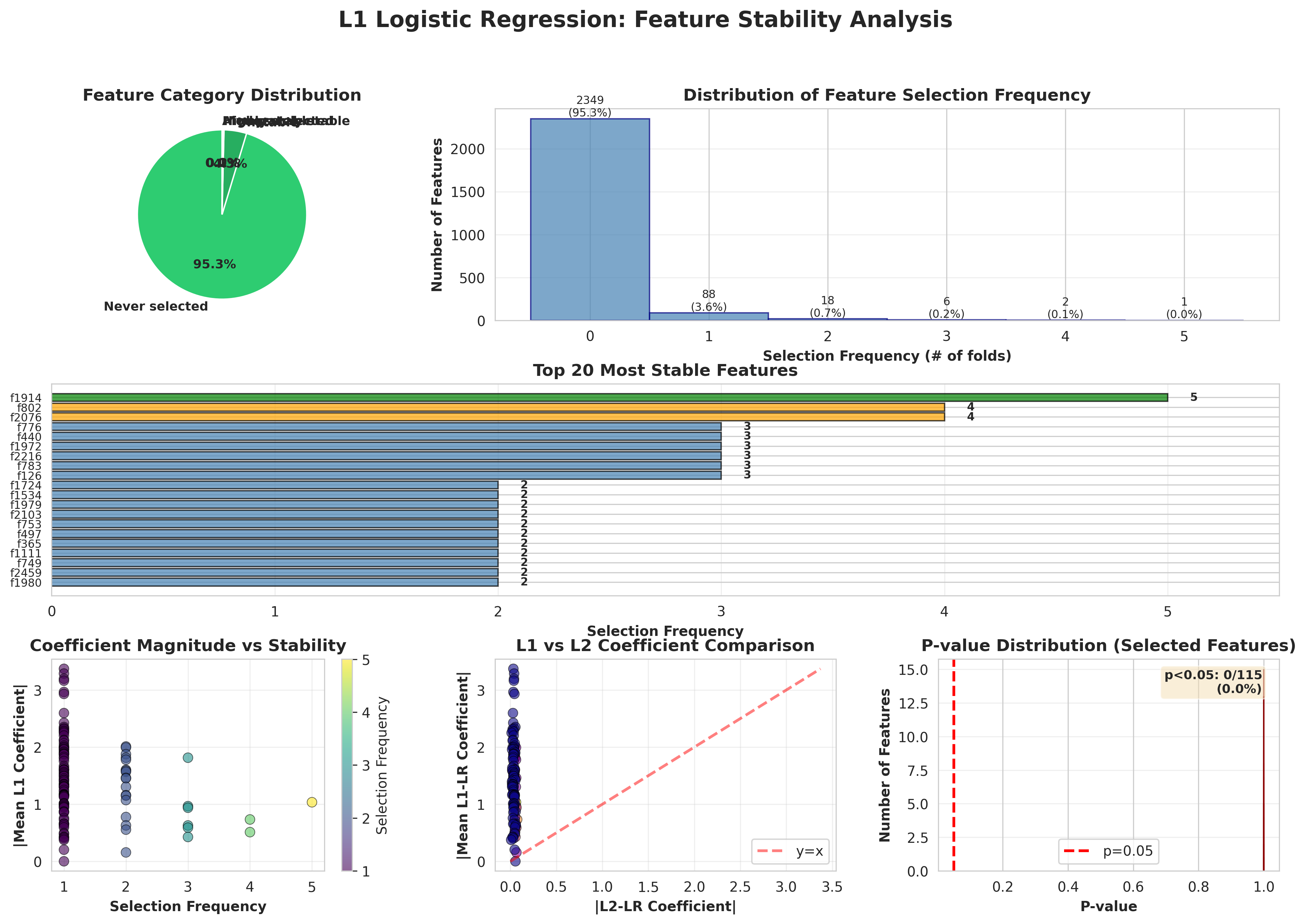}
    \caption{Stability Overview of L1 Regression (Threshold 0.6)}
    \label{fig:l1_lr_stability_06}
\end{figure*}

\textbf{Critical observation:} Every single always-selected feature across all thresholds has Standard LR coefficient |$\beta$|<0.07 and p-value >0.999, indicating no statistical significance. These features are consistently selected by L1 not because they are important, but because they are arbitrary representatives picked first from correlated question clusters during L1's optimization path. Figure~\ref{fig:l1_lr_stability_03} through Figure~\ref{fig:l1_lr_stability_06} visualize the selection patterns across thresholds.

\subpara{Per-Fold Performance Details}
\label{app:l1_lr_fold_details}
Table~\ref{tab:l1_lr_fold_breakdown} shows detailed per-fold statistics revealing the relationship between feature counts, hyperparameters, and performance.

\begin{table*}[h]
\centering
\caption{L1 LR Per-Fold Performance Breakdown}
\label{tab:l1_lr_fold_breakdown}
\begin{tabular}{clrrrrr}
\toprule
\textbf{Threshold} & \textbf{Fold} & \textbf{F1} & \textbf{Acc} & \textbf{\# Feat} & \textbf{C} & \textbf{Freq} \\
\midrule
\multirow{5}{*}{0.3}
& 1 & 0.708 & 0.761 & 362 & 100.00 & 0.155±0.147 \\
& 2 & 0.756 & 0.777 & 35 & 0.17 & 0.023±0.072 \\
& 3 & 0.763 & 0.790 & 85 & 2.15 & 0.059±0.095 \\
& 4 & 0.734 & 0.781 & 289 & 100.00 & 0.144±0.138 \\
& 5 & 0.718 & 0.768 & 288 & 100.00 & 0.144±0.135 \\
\midrule
\multirow{5}{*}{0.4}
& 1 & 0.705 & 0.752 & 192 & 100.00 & 0.155±0.147 \\
& 2 & 0.717 & 0.744 & 21 & 0.17 & 0.023±0.072 \\
& 3 & 0.740 & 0.765 & 44 & 2.15 & 0.059±0.095 \\
& 4 & 0.696 & 0.734 & 169 & 100.00 & 0.144±0.138 \\
& 5 & 0.755 & 0.793 & 139 & 100.00 & 0.144±0.135 \\
\midrule
\multirow{5}{*}{0.5}
& 1 & 0.724 & 0.761 & 112 & 100.00 & 0.155±0.147 \\
& 2 & 0.671 & 0.698 & 13 & 0.17 & 0.023±0.072 \\
& 3 & 0.688 & 0.710 & 19 & 2.15 & 0.059±0.095 \\
& 4 & 0.741 & 0.768 & 83 & 100.00 & 0.144±0.138 \\
& 5 & 0.823 & 0.848 & 82 & 100.00 & 0.144±0.135 \\
\midrule
\multirow{5}{*}{0.6}
& 1 & 0.743 & 0.773 & 65 & 100.00 & 0.155±0.147 \\
& 2 & 0.623 & 0.655 & 8 & 0.17 & 0.023±0.072 \\
& 3 & 0.713 & 0.735 & 14 & 2.15 & 0.059±0.095 \\
& 4 & 0.676 & 0.705 & 34 & 100.00 & 0.144±0.138 \\
& 5 & 0.749 & 0.781 & 35 & 100.00 & 0.144±0.135 \\
\bottomrule
\end{tabular}
\end{table*}

Key observations emerge from this analysis. Fold 2 consistently has the lowest C (0.17), selecting fewest features (8 to 35), often with worst performance. Folds 1, 4, and 5 have highest C (100), selecting most features (34 to 362), with varied performance. Despite 10× feature count variation within thresholds, F1 varies only 6 to 20 points. Threshold 0.5 shows extreme variance: best fold (F1=0.823) and some of worst (F1=0.671). No consistent relationship exists between number of features and performance within or across thresholds.

\subpara{Negative Correlation Analysis}
The correlation between L1 and Standard LR coefficients becomes increasingly negative as thresholds tighten. At threshold 0.3, correlation is r = $-$0.026 (essentially zero). At threshold 0.4, r = $-$0.059 (weak negative). At threshold 0.5, r = $-$0.106 (moderate negative). At threshold 0.6, r = $-$0.249 (strong negative). This monotonic deterioration proves that as L1 is forced to select fewer features, it increasingly prioritizes features that Standard LR considers less important. By threshold 0.6, L1 and Standard LR are in substantial disagreement, demonstrating that aggressive sparsity causes L1 to make systematically poor selections.

\paragraph{L1 SVC Detailed Results}
\label{app:l1_svc_results}

\subpara{Complete Selection Statistics}
Table~\ref{tab:l1_svc_full_stats} presents comprehensive statistics for L1 SVC across threshold configurations.

\begin{table*}[h]
\centering
\caption{Complete L1 SVC Selection Statistics}
\label{tab:l1_svc_full_stats}
\begin{tabular}{lrrrr}
\toprule
\textbf{Metric} & \textbf{0.3} & \textbf{0.4} & \textbf{0.5} & \textbf{0.6} \\
\midrule
\multicolumn{5}{l}{\textit{Selection Consistency Across Folds}} \\
Always (5/5 folds) & 386 (15.7\%) & 158 (6.4\%) & 64 (2.6\%) & 24 (1.0\%) \\
Highly stable (4/5) & 380 (15.4\%) & 170 (6.9\%) & 89 (3.6\%) & 37 (1.5\%) \\
Moderately stable (3/5) & 475 (19.3\%) & 327 (13.3\%) & 179 (7.3\%) & 99 (4.0\%) \\
Unstable (1-2/5) & 1222 (49.6\%) & 1729 (70.2\%) & 1516 (61.5\%) & 928 (37.7\%) \\
Never selected & 1 (0.04\%) & 80 (3.2\%) & 616 (25.0\%) & 1376 (55.8\%) \\
\midrule
\multicolumn{5}{l}{\textit{Performance Metrics}} \\
Mean F1 & 0.769±0.015 & 0.746±0.027 & 0.744±0.020 & 0.751±0.032 \\
Mean Accuracy & 0.805±0.011 & 0.785±0.018 & 0.781±0.015 & 0.786±0.024 \\
\midrule
\multicolumn{5}{l}{\textit{Per-Fold Feature Count Variation}} \\
Mean \# features & 1340.6 & 791.6 & 564.0 & 341.6 \\
Std \# features & 846.3 & 746.4 & 649.9 & 369.0 \\
Min features & 1031 & 578 & 311 & 149 \\
Max features & 2463 & 2369 & 1772 & 974 \\
Range (max/min) & 2.4× & 4.1× & 5.7× & 6.5× \\
\midrule
\multicolumn{5}{l}{\textit{Hyperparameter Statistics}} \\
C range & 27.8-100 & 27.8-100 & 27.8-100 & 27.8-100 \\
C ratio (max/min) & 3.6× & 3.6× & 3.6× & 3.6× \\
\bottomrule
\end{tabular}
\end{table*}

Comparison with L1 LR reveals key differences. L1 SVC achieves 32× more stable selections (158 vs. 5 always-selected at threshold 0.4), but selects 7 to 23× more features per fold (average 792 vs. 113 features at threshold 0.4). It shows better performance: 0.746 vs. 0.722 F1 at threshold 0.4 (2.4 point gain). L1 SVC exhibits less hyperparameter sensitivity with C varying only 3.6× vs. 599× for L1 LR, and smaller feature count variation with 2.4 to 6.5× range vs. 8 to 10× for L1 LR.

\begin{figure*}
    \centering
    \includegraphics[width=0.8\linewidth]{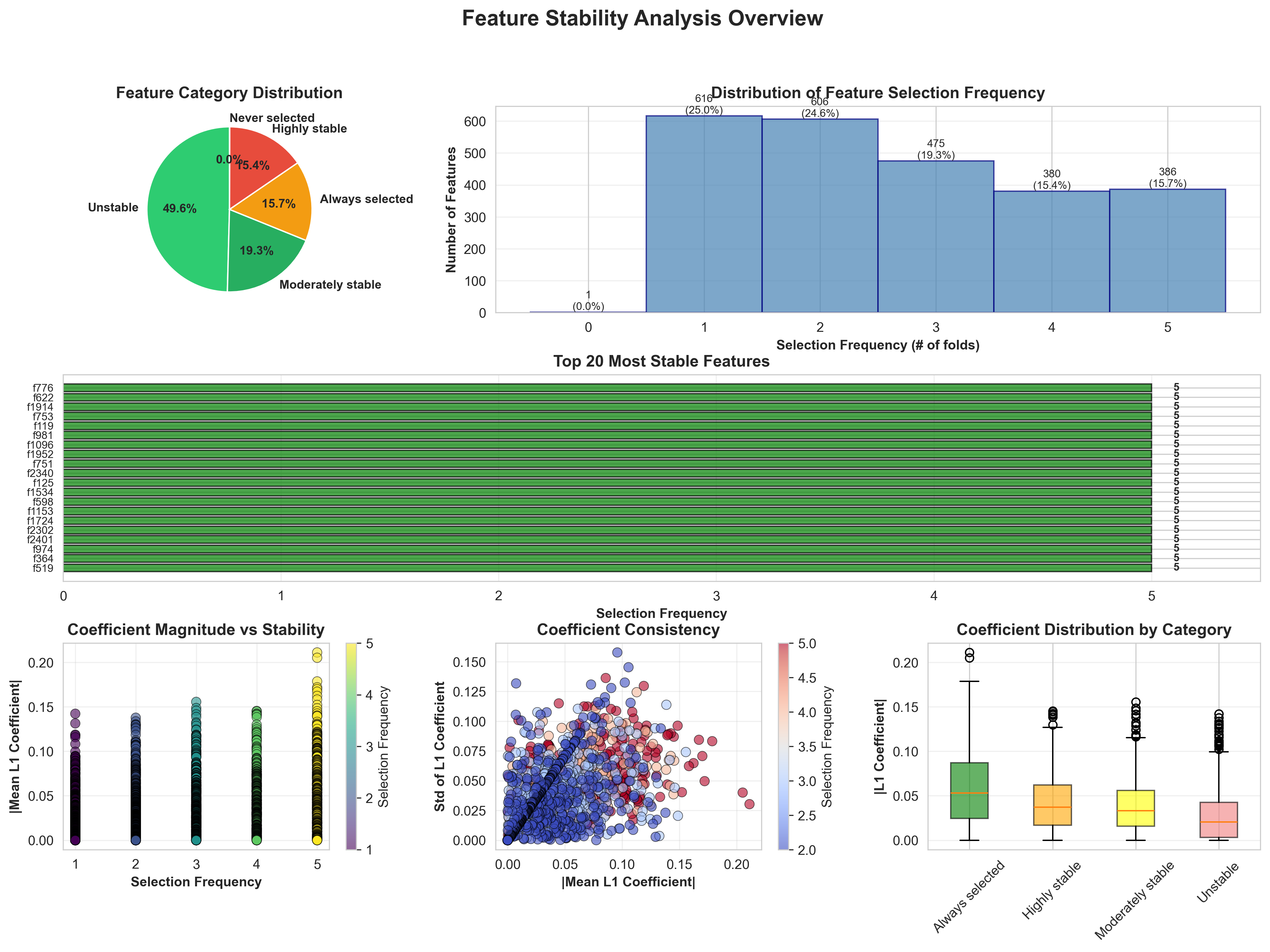}
    \caption{Stability Overview of L1 SVC (Threshold 0.3)}
    \label{fig:l1_svc_stability_03}
\end{figure*}

\begin{figure*}
    \centering
    \includegraphics[width=0.8\linewidth]{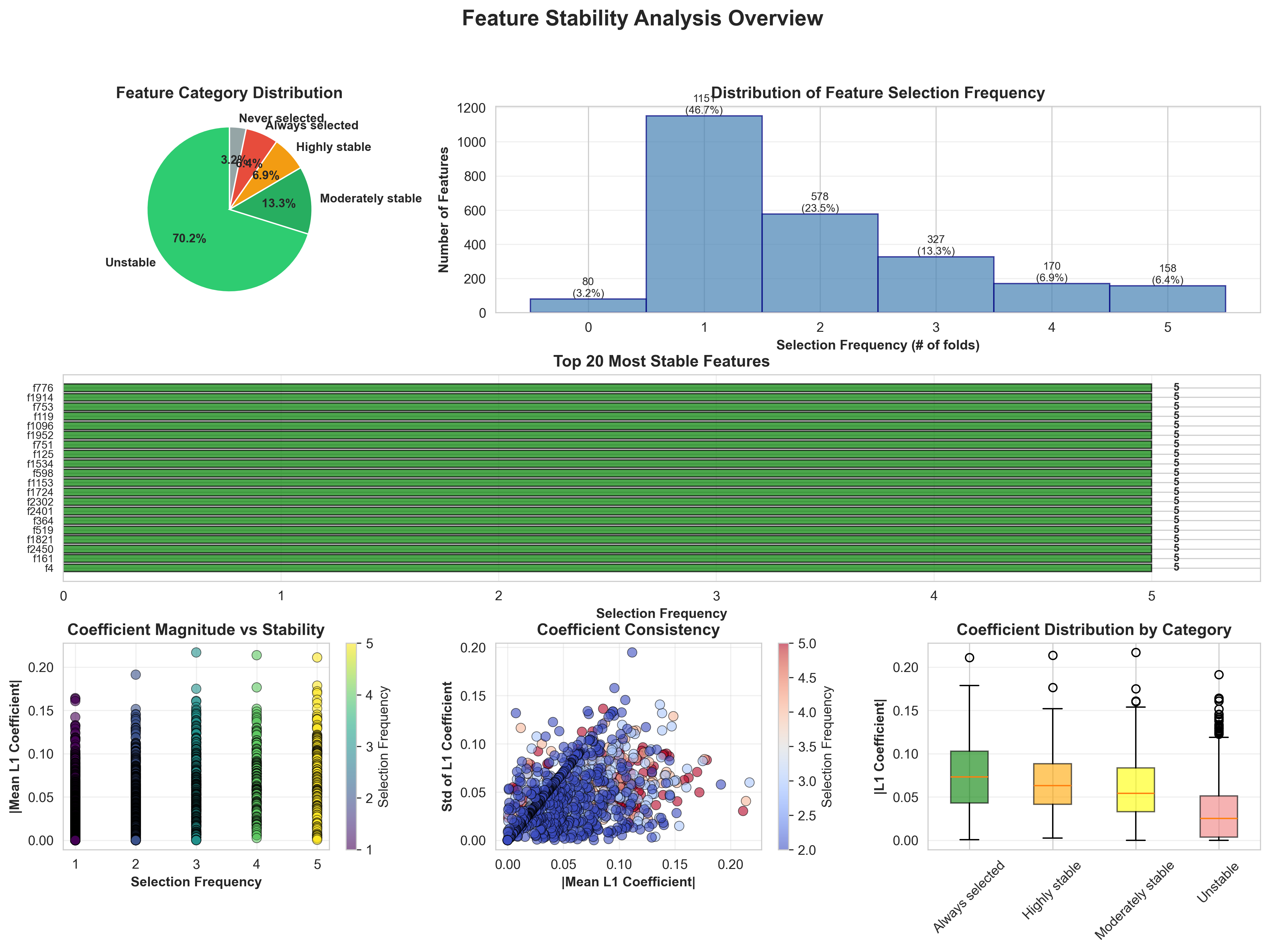}
    \caption{Stability Overview of L1 SVC (Threshold 0.4)}
    \label{fig:l1_svc_stability_04}
\end{figure*}

\begin{figure*}
    \centering
    \includegraphics[width=0.8\linewidth]{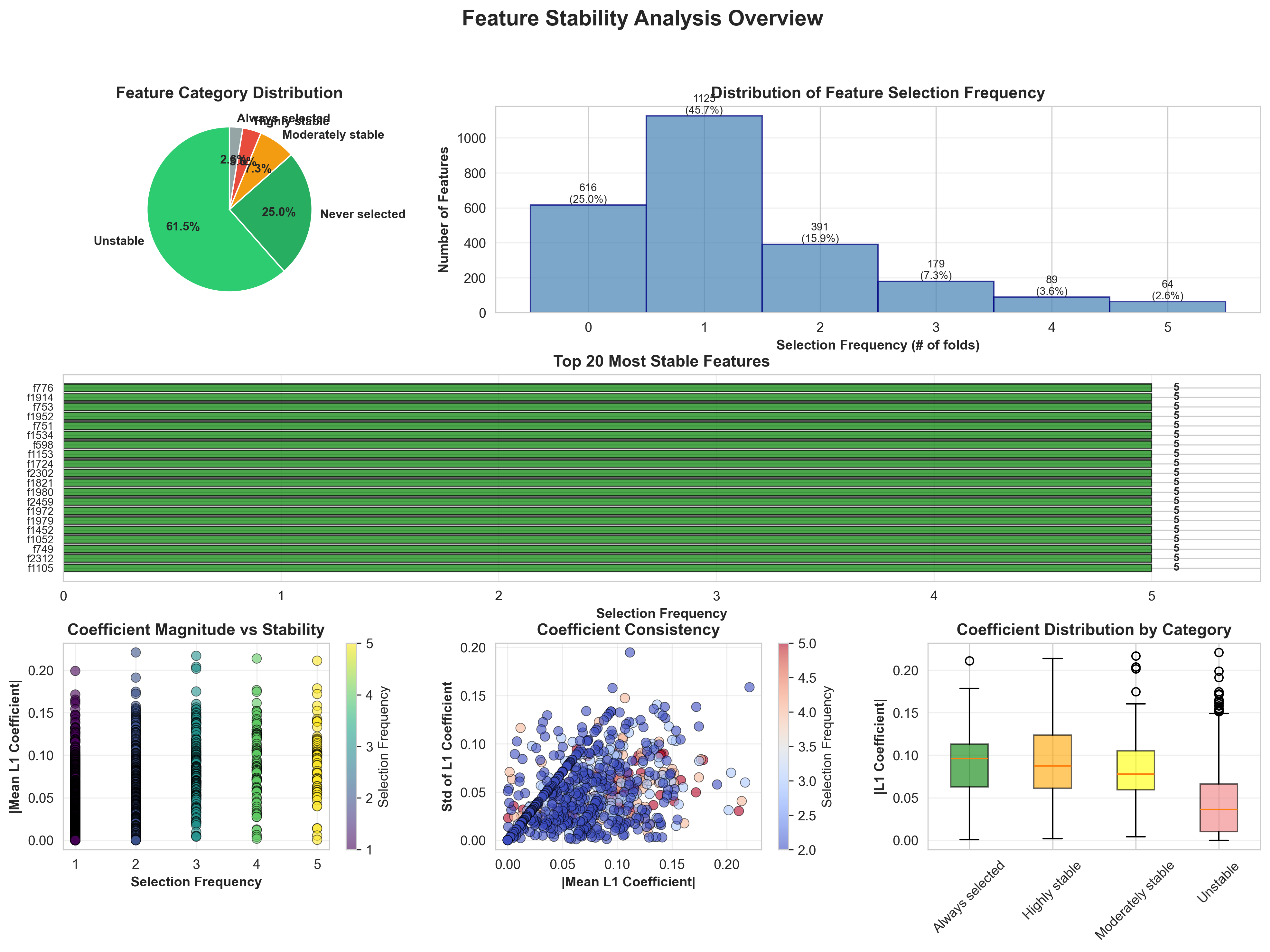}
    \caption{Stability Overview of L1 SVC (Threshold 0.5)}
    \label{fig:l1_svc_stability_05}
\end{figure*}

\begin{figure*}
    \centering
    \includegraphics[width=0.8\linewidth]{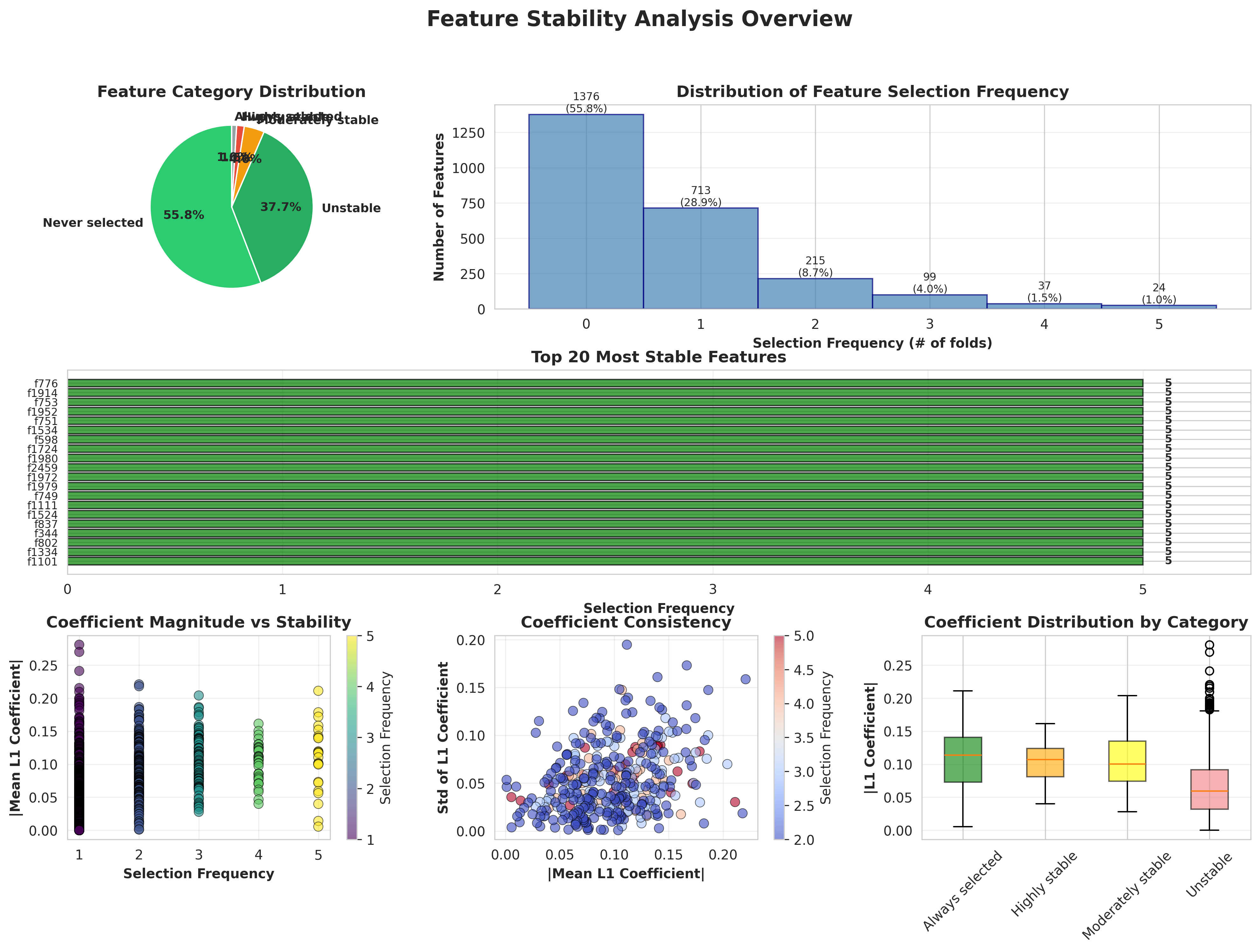}
    \caption{Stability Overview of L1 SVC (Threshold 0.6)}
    \label{fig:l1_svc_stability_06}
\end{figure*}

\subpara{Always-Selected Features Analysis}
Unlike L1 LR, L1 SVC's always-selected features include genuinely high-weighted features. However, the set changes dramatically with threshold. At threshold 0.3, 386 always-selected features (15.7\% of total) emerge. The top 10 include f776 ($\beta$=$-$0.211), f622 ($\beta$=0.205), f1914 ($\beta$=0.179), f753 ($\beta$=0.172), and f119 ($\beta$=0.170). These are spread across many legal reasoning categories and represent comprehensive retention rather than sparse selection. At threshold 0.6, only 24 always-selected features (1.0\% of total) remain. The top 10 include f776 ($\beta$=$-$0.211), f1914 ($\beta$=0.179), f753 ($\beta$=0.172), f1952 ($\beta$=0.158), and f751 ($\beta$=$-$0.152). These overlap with threshold 0.3 top features but represent a dramatically reduced set, still more stable than L1 LR's single always-selected feature. Figure~\ref{fig:l1_svc_stability_03} through Figure~\ref{fig:l1_svc_stability_06} illustrate these patterns.

\subpara{Per-Fold Performance Details}
Table~\ref{tab:l1_svc_fold_breakdown} shows L1 SVC's per-fold statistics.

\begin{table*}[h]
\centering
\caption{L1 SVC Per-Fold Performance Breakdown}
\label{tab:l1_svc_fold_breakdown}
\begin{tabular}{clrrrrr}
\toprule
\textbf{Threshold} & \textbf{Fold} & \textbf{F1} & \textbf{Acc} & \textbf{\# Feat} & \textbf{C} & \textbf{Freq} \\
\midrule
\multirow{5}{*}{0.4}
& 1 & 0.702 & 0.756 & 610 & 27.83 & 0.311±0.172 \\
& 2 & 0.785 & 0.811 & 2369 & 100.00 & 0.579±0.128 \\
& 3 & 0.743 & 0.777 & 597 & 27.83 & 0.307±0.168 \\
& 4 & 0.753 & 0.789 & 604 & 27.83 & 0.309±0.163 \\
& 5 & 0.749 & 0.789 & 578 & 27.83 & 0.306±0.161 \\
\bottomrule
\end{tabular}
\end{table*}

A key observation emerges: Fold 2 with C=100 selects 2369 features (96\% of all features) and achieves best performance (F1=0.785). Other folds with C=27.83 select approximately 600 features with lower but stable performance (F1=0.702 to 0.753). This demonstrates L1 SVC's tendency toward dense solutions when C is high, contradicting the sparsity objective of L1 regularization.

\paragraph{Cross-Method Comparison}
\label{app:l1_cross_method}

\subpara{Selection Overlap Analysis}
We analyze which features are selected by both L1 LR and L1 SVC at threshold 0.4. L1 LR has 5 always-selected features, while L1 SVC has 158. The overlap (always-selected in both) contains only 3 features, representing 60\% of L1 LR's selections but just 1.9\% of L1 SVC's. When considering features selected in at least 3 folds, L1 LR has 29 such features while L1 SVC has 655. The overlap in this category contains 11 features, representing 38\% of L1 LR but only 1.7\% of L1 SVC. This low overlap (38 to 60\%) proves that different L1-based methods identify different "important" features, further demonstrating selection instability. If a true essential question set existed, both methods should converge to it.

\subpara{Why Does L1 SVC Select More Features}
The key difference stems from the optimization objectives. L1 Logistic Regression minimizes 
$$\min_{\mathbf{w}, b} \left\{ -\frac{1}{n}\sum_{i=1}^{n} \log P(y_i | \mathbf{x}_i; \mathbf{w}, b) + \lambda \|\mathbf{w}\|_1 \right\}$$
The log-loss is smooth and convex, with L1 regularization pushing many weights to exactly zero when C is small. In contrast, L1 SVC minimizes
$$\min_{\mathbf{w}, b} \left\{ \frac{1}{n}\sum_{i=1}^{n} \max(0, 1 - y_i(\mathbf{w}^\top \mathbf{x}_i + b)) + \lambda \|\mathbf{w}\|_1 \right\}$$
The hinge loss has a non-smooth "hinge" at margin boundaries. For our high-dimensional, correlated feature space, SVC's dual formulation tends to utilize many support vectors, each requiring non-zero weights for its associated features. This dual structure makes L1 SVC less aggressive at zeroing weights than L1 LR. When C is large (weak regularization), L1 SVC approaches unregularized SVC, which uses many features to maximize margin. This explains Fold 2's selection of 2369/2464 features.

\paragraph{Interpretation and Implications}
Three fundamental reasons explain the absence of a universal essential question set.

\subpara{Why No Stable Essential Set Exists} \textit{Question generation process creates context-specific, not universal, questions.} Our reasoning questions were generated for specific case-issue pairs (approximately 300 pairs, approximately 8 questions each = 2464 total). These questions are contextualized to particular legal scenarios rather than being a universal reasoning bank. For example, "Is the issue central to the insurer's stated reason for denying the claim?" only applies to insurance cases, "Does the employee's conduct fall under the protected activity definition?" only applies to employment cases, and "Was proper notice given according to the lease terms?" only applies to property/contract cases. A question that is critical for one case type is noise for others. This domain-specificity means no fixed subset works universally; effective reasoning requires context-dependent weighting. \textit{Massive redundancy from correlated questions.} With 2464 questions, many are semantically similar due to rephrasing of the same concept ("Is this a legal vs. factual question?" versus "Does answering require applying law?"), different levels of abstraction ("Is this jurisdictional?" versus "Does this court have authority?" versus "Is venue proper?"), and domain-specific variants of general questions. This redundancy means many different sparse subsets capture similar information, explaining why L1 achieves 0.70 to 0.75 F1 with completely different feature sets across folds. Any reasonable subset of approximately 50 to 200 questions suffices because they collectively cover the key reasoning dimensions, but which specific 50 to 200 questions doesn't matter much. 

\subpara{Linear models enable implicit contextualization.} Standard linear models achieve 80\% F1 (versus L1's 70 to 75\%) by retaining all questions with learned weights. The linear combination $\mathbf{w}^\top \mathbf{f} = \sum_{j=1}^{h} w_j f_j$ enables domain-specific questions to contribute conditionally. When feature $f_j$ (domain-specific) co-occurs with features indicating that domain (e.g., insurance keywords in case facts), their combined contribution is large. When domain indicators are absent, $f_j$'s contribution is attenuated by negative contributions from other features. This implicit contextualization through weighted linear combinations is more sophisticated than L1's binary keep/discard decisions, explaining the 5 to 10 point performance gap.

\subpara{Implications for Legal AI Systems}
The findings have practical implications for building legal AI systems.

\textit{Favor retention over selection.} Systems should retain comprehensive question sets with learned weights rather than attempting to identify "essential" subsets. The 5 to 10 point F1 gap between Standard linear models (80\%) and L1 methods (70 to 75\%) quantifies the cost of feature elimination in correlation-aware classification tasks.

\textit{Contextualization is key.} Legal reasoning inherently requires context-dependent weighting. Domain-specific questions that appear noisy globally (low marginal contribution) are critical in specialized contexts. Systems must preserve these questions and adjust their influence based on case characteristics.

\textit{Interpretability versus performance tradeoffs.} L1 provides sparse, seemingly interpretable models (e.g., "only 50 questions matter"), but this interpretability is illusory. Selected questions are unstable (vary across folds), different methods select different questions, selected questions have no special importance (near-zero Standard LR weights), and performance suffers due to elimination of contextual questions. True interpretability requires analyzing Standard linear model weights to understand which reasoning aspects receive highest weights across all questions, not which arbitrary subset L1 happens to select.

\subpara{Summary}
The comprehensive stability analysis across L1 LR and L1 SVC demonstrates five key findings. First, no stable universal essential question set exists for legal issue relevance. Second, L1-based selection achieves competitive performance with unstable, method-dependent feature subsets. Third, multiple sparse/dense combinations work equivalently due to massive feature redundancy. Fourth, retention with adaptive weighting (Standard models: 80\% F1) outperforms elimination (L1: 70 to 75\% F1). Fifth, contextualization through linear combinations is superior to binary feature selection. These findings conclusively validate Challenge 2 and justify our correlation-aware retention-based approach over sparsity-seeking feature selection methods.

\paragraph{Feature Stability Visualizations}
We illustrate the feature selection patterns through stability overview figures for both methods across all threshold configurations.

\onecolumn
\begin{center}
\footnotesize
\begin{longtable}{clp{4.5cm}p{5.5cm}}
\caption{Top 25 Positively and Top 25 Negatively Weighted Reasoning Questions selected from the \ourname\ linear model for the usability study.}
\label{tab:reasoning_questions}\\

\toprule
\textbf{\#} & \textbf{Weight} & \textbf{Reasoning Question} & \textbf{Rationale} \\
\midrule
\endfirsthead

\multicolumn{4}{l}{\textit{Table~\ref{tab:reasoning_questions} continued.}}\\
\toprule
\textbf{\#} & \textbf{Weight} & \textbf{Reasoning Question} & \textbf{Rationale} \\
\midrule
\endhead

\midrule
\multicolumn{4}{r}{\textit{Continued on next page.}}\\
\endfoot

\bottomrule
\endlastfoot

\multicolumn{4}{l}{\textit{Top 25 Positively Weighted Questions (indicative of relevance)}} \\
\midrule
1  & $+$ & Has the court explicitly stated that the issue would not affect its determination on the merits of the case? & When a court explicitly states that an issue does not impact its decision on the merits, it often indicates the issue's irrelevance to the core dispute. \\
2  & $+$ & Does the issue address the core legal question raised by the parties' actions? & The relevance of an issue is often determined by its relationship to the fundamental legal questions arising from the parties' actions and claims. \\
3  & $+$ & Does the issue address the main point of contention in the dispute as described in the facts? & This confirms whether the issue captures the essence of the legal controversy, which is crucial for establishing its relevance to the case. \\
4  & $+$ & Does the issue require an examination of the specific terms or conditions of the agreement in light of the alleged illegality? & This determines if the issue necessitates a detailed analysis of the agreement's terms in relation to illegal activities, often central to determining contract validity. \\
5  & $+$ & Did the court indicate that this issue would not affect the determination on the merits of the case? & When a court explicitly states that an issue does not impact the case's outcome, it often suggests the issue is not central to the main dispute. \\
6  & $+$ & Is the issue central to determining the legality or propriety of the proceedings described in the facts? & This establishes whether the issue is fundamental to assessing the overall validity of the legal process. \\
7  & $+$ & Does the issue address the key elements that led to the court's finding in the scenario? & Evaluating whether the issue relates to the court's reasoning helps determine its significance to the case's outcome. \\
8  & $+$ & Is the issue focused on a specific legal conclusion rather than a general legal principle? & Specific legal conclusions tied to the case are more likely to be directly relevant than general legal principles. \\
9  & $+$ & Is the issue related to the legality or enforceability of an agreement mentioned in the facts? & Assessing the legal validity of an agreement is often at the core of contractual disputes. \\
10 & $+$ & Does the issue address a key point of contention that was ruled upon by the courts mentioned in the facts? & This checks whether the issue aligns with the core legal matters judicially considered. \\
11 & $+$ & Does the issue relate to the legality or enforceability of the agreement described in the facts? & This checks whether the issue touches on the legal validity of the agreement, often a key factor in determining relevance. \\
12 & $+$ & Does the issue address the core dispute that led to legal proceedings? & This determines whether the issue is at the heart of the conflict that resulted in the case. \\
13 & $+$ & Does the issue pertain to a matter that could affect the rights or obligations of the parties involved? & This assesses whether the issue has practical implications for the parties, crucial for establishing relevance. \\
14 & $+$ & Is the issue related to the primary action or counteraction taken by the parties in the scenario? & This establishes whether the issue is connected to the main moves or decisions made by the parties. \\
15 & $+$ & Does the issue directly address the primary legal dispute in the scenario? & The relevance of an issue depends on whether it addresses the core legal dispute rather than peripheral matters. \\
16 & $+$ & Does the issue involve a decision made by the highest court mentioned in the scenario? & Whether the highest court's decision is involved often represents the final word on the matter. \\
17 & $+$ & Is the issue focused on a specific legal remedy or action that could be taken by the court? & This assesses whether the issue involves a concrete legal action, often indicating direct relevance. \\
18 & $+$ & Is the issue related to a potential legal violation or illegal act mentioned in the facts? & Issues involving matters of legality are often central to determining the outcome of a case. \\
19 & $+$ & Is the issue related to a specific legal claim or defense mentioned in the facts? & Issues pertaining to specific legal claims or defenses raised by the parties are often highly relevant. \\
20 & $+$ & Is the issue directly connected to the outcome or judgment described in the scenario? & This assesses whether the issue is linked to the consequences or rulings that resulted from the actions in question. \\
21 & $+$ & Does the issue pertain to a matter that is no longer in contention in the current legal proceeding? & Issues no longer actively disputed may be less relevant to the current stage of proceedings. \\
22 & $+$ & Is the issue related to a specific clause or condition that is being contested by the parties? & An issue focused on a specific, contested clause is more likely to be central to the dispute. \\
23 & $+$ & Does resolving this issue potentially affect the rights and liabilities of all parties involved? & An issue impacting the rights and liabilities of all parties is likely at the heart of the legal controversy. \\
24 & $+$ & Is the issue framed in a way that directly challenges or supports the main finding described in the scenario? & Aligning the issue with the central finding reinforces its relevance to the core matter. \\
25 & $+$ & Does resolving this issue directly impact the outcome of the disciplinary or legal proceedings? & Issues with a direct bearing on the case's outcome are typically highly relevant. \\
\midrule
\multicolumn{4}{l}{\textit{Top 25 Negatively Weighted Questions (indicative of irrelevance)}} \\
\midrule
26 & $-$ & Has the issue already been conclusively decided by a higher court in the scenario? & The finality of a higher court's decision can affect the relevance of an issue in ongoing proceedings. \\
27 & $-$ & Does resolving this issue definitively settle the legal dispute between the parties? & An issue's ability to conclusively resolve the dispute is a strong indicator of its relevance. \\
28 & $-$ & Is there a direct financial consequence mentioned in relation to the issue? & Financial implications often indicate that an issue is material rather than incidental. \\
29 & $-$ & Does the issue relate to the legal process or procedural aspects of the case? & Procedural matters can be significant, but may not address the substantive coverage. \\
30 & $-$ & Is the issue specifically mentioned in the court's reasoning for its decision? & Courts' explicit reasoning often indicates which issues were central to their decisions. \\
31 & $-$ & Is the issue related to the insurance companies' reason for denying the claim? & The primary controversy revolves around the denial reason, not the documentation process. \\
32 & $-$ & Is the issue related to the financial obligations outlined in the agreement? & Financial obligations are frequently at the core of contractual disputes. \\
33 & $-$ & Is the issue related to a specific professional standard or ethical requirement mentioned in the facts? & Connecting the issue to professional standards helps establish relevance in professional misconduct cases. \\
34 & $-$ & Does the issue relate to the interpretation of evidence presented in the case? & Issues involving evidence interpretation are typically relevant to the central matters of a dispute. \\
35 & $-$ & Is the issue connected to the economic impact of the terms described in the scenario? & This explores whether the issue has direct financial implications related to the facts. \\
36 & $-$ & Is the issue rendered moot by subsequent developments in the case? & An issue may become irrelevant if superseded or rendered moot by later rulings. \\
37 & $-$ & Is the issue about interpreting a specific clause in the agreement? & Understanding whether the issue involves contract interpretation is crucial for assessing relevance. \\
38 & $-$ & Does the issue concern the handling of funds in a professional capacity? & This focuses on whether the issue involves financial responsibilities, often central to professional conduct cases. \\
39 & $-$ & Is the issue related to the interpretation of a specific contractual term? & Contractual interpretation issues are often core to legal disputes. \\
40 & $-$ & Is the issue a fundamental legal principle that applies to all similar cases? & Fundamental legal principles may not always be directly relevant to the specific controversy. \\
41 & $-$ & Does the issue involve a challenge to the enforceability of a contractual provision? & Challenges to enforceability are typically central to contract disputes. \\
42 & $-$ & Is the issue connected to the economic impact of the contract terms on the parties? & Economic consequences of contract terms often form the heart of contractual disputes. \\
43 & $-$ & Is the issue tied to a specific request for relief mentioned in the facts? & Connection to requested relief suggests the issue is central to resolving the dispute. \\
44 & $-$ & Does the issue involve the interpretation or application of a relevant law or regulation? & Assessing whether the issue involves legal interpretation helps determine its significance. \\
45 & $-$ & Is the issue about interpreting a specific clause in the contract? & Whether the issue focuses on contract interpretation is crucial for assessing its relevance. \\
46 & $-$ & Is the issue tied to the legal standard or rule applied in the case? & Linking the issue to the applicable legal standard helps establish its significance. \\
47 & $-$ & Does the issue directly address the enforcement of a contractual agreement mentioned in the facts? & This establishes whether the issue is directly related to the core contractual dispute. \\
48 & $-$ & Is the issue related to the main financial transaction or arrangement in dispute? & Connecting the issue to the primary financial matter helps establish its centrality. \\
49 & $-$ & Does resolving this issue alone determine the outcome of the legal dispute? & An issue's ability to singularly resolve the case is a key factor in assessing legal relevance. \\
50 & $-$ & Is the issue focused on interpreting specific terms of the agreement in question? & Issues centered on interpreting specific contractual terms are typically more relevant than abstract principles. \\

\end{longtable}
\end{center}
\twocolumn

\end{document}